\documentclass[lettersize,journal]{IEEEtran}
\usepackage{amsmath,amsfonts}
\usepackage{algorithmic}
\usepackage{algorithm}
\usepackage{array}
\usepackage[caption=false,font=normalsize,labelfont=sf,textfont=sf]{subfig}

\usepackage{textcomp}
\usepackage{stfloats}
\usepackage{url}
\usepackage{verbatim}
\usepackage{graphicx}
\usepackage{cite}
\usepackage{amssymb}
\usepackage{mathrsfs}
\usepackage{multirow}
\usepackage{float}
\usepackage{graphicx}

\usepackage{textcase} 

\usepackage{makecell}
\usepackage[font=footnotesize,labelfont=bf]{caption}
\usepackage{subcaption}
\usepackage{rotating}
\usepackage{color}

\usepackage{afterpage}
\usepackage{colortbl}
\usepackage{xcolor}
\colorlet{light}{green!50}
\colorlet{lightlight}{green!25}

\colorlet{colorSep}{blue!5}  
\usepackage{mathtools}
\usepackage{tikz}

\usepackage[colorlinks,
            linkcolor=blue,
            anchorcolor=blue,
            citecolor=blue]{hyperref}

\newcommand{\FrameworkNM}[1]{SPORTS#1}

\newcommand*\circled[1]{\tikz[baseline=(char.base)]{
            \node[shape=circle,draw,inner sep=0.1pt] (char) {#1};}}

\begin{document}

\title{
\FrameworkNM{}: Simultaneous Panoptic Odometry, Rendering, Tracking and Segmentation for Urban Scenes Understanding
}

\author{
Zhiliu Yang,~\IEEEmembership{Member,~IEEE,}
Jinyu Dai,~\IEEEmembership{Student Member,~IEEE,}
Jianyuan Zhang,~\IEEEmembership{Student Member,~IEEE,}\\
Zhu Yang,~\IEEEmembership{Member,~IEEE}

\thanks{
This work was supported in part by Yunnan Science Foundation of Yunnan Provincial Department of Science and Technology, China under Grant 202301AU070200, and in part by the Program of Yunnan Key Laboratory of Intelligent Systems and Computing under Grant 202405AV340009.}

\thanks{
Zhiliu Yang, Jinyu Dai, and Jianyuan Zhang are with the School of Information Science and Engineering, Yunnan University, Kunming, Yunnan 650500, China.
}

\thanks{
Zhu Yang is with the School of Information and Electronics, Beijing Institute of Technology, Beijing 100081, China.}

\thanks{
Zhiliu Yang is also with the Yunnan Key Laboratory of Intelligent Systems and Computing, Yunnan University, Kunming, Yunnan 650500, China.
}

\thanks{
(Corresponding authors: Zhiliu Yang; Zhu Yang.)}

\thanks{
This manuscript is the accepted version of IEEE Transactions on Multimedia. ~\copyright~2025 IEEE. Personal use of this material is permitted. Permission from IEEE must be obtained for all other uses, in any current or future media, including reprinting/republishing this material for advertising or promotional purposes, creating new collective works, for resale or redistribution to servers or lists, or reuse of any copyrighted component of this work in other works.}}

\markboth{Journal of \LaTeX\ Class Files,~Vol.~14, No.~8, August~2025}%
{Shell \MakeLowercase{\textit{et al.}}: A Sample Article Using IEEEtran.cls for IEEE Journals}

\IEEEpubid{0000--0000/00\$00.00~\copyright~2025 IEEE }

\maketitle

\begin{abstract}
The scene perception, understanding, and simulation are fundamental techniques for embodied-AI agents, while existing solutions are still prone to segmentation deficiency, dynamic objects' interference, sensor data sparsity, and view-limitation problems. This paper proposes a novel framework, named \FrameworkNM{}, for holistic scene understanding via tightly integrating Video Panoptic Segmentation (VPS), Visual Odometry (VO), and Scene Rendering (SR) tasks into an iterative and unified perspective. Firstly, VPS designs an adaptive attention-based geometric fusion mechanism to align cross-frame features via enrolling the pose, depth, and optical flow modality, which automatically adjust feature maps for different decoding stages. And a post-matching strategy is integrated to improve identities tracking. In VO, panoptic segmentation results from VPS are combined with the optical flow map to improve the confidence estimation of dynamic objects, which enhances the accuracy of the camera pose estimation and completeness of the depth map generation via the learning-based paradigm. Furthermore, the point-based rendering of SR is beneficial from VO, transforming sparse point clouds into neural fields to synthesize high-fidelity RGB views and twin panoptic views. Extensive experiments on three public datasets demonstrate that our attention-based feature fusion outperforms most existing state-of-the-art methods on the odometry, tracking, segmentation, and novel view synthesis tasks.
\end{abstract}


\section{Introduction}
\label{sec::introduce}

\IEEEPARstart{W}{ith} the foreseeable future of the ubiquitous deployment of embodied-AI agents, such as self-driving vehicles \cite{9782567,xie2023s, turki2023suds, li2024dgnr}, quadruped robots \cite{8793865,10610561} or humanoid robots \cite{10610948}, a holistic urban scene understanding ability is more and more crucial for agents to physically fulfill tasks like perception \cite{10499836,9447924}, localization \cite{zhang2021geometry}, and collision avoidance \cite{9782567}.
Besides, the holistic scene understanding can be exploited to create a digital twin of the urban environment to mitigate the reality gap, it acts as a simulation platform for agents to learn and verify the unexpected corner cases, to enhance the safety of autonomous driving at a lower cost \cite{li2023read,zhou2024drivinggaussian, yan2024street}. Unfortunately, precisely perceiving and modeling urban scenes are challenging due to various object sizes, complicated geometry details, and varying road conditions etc. \cite{9782567, turki2023suds, li2024dgnr}, and the scene is usually depicted via limited views and sparse data \cite{zhou2024drivinggaussian}.

Existing methods such as \cite{kim2020simvodis, ji2021full, mumuni2022deep} have started to tackle the \textit{information silos problem} of holistic scene understanding. They leverage mutual benefits of different modalities, such as optical flow, depth and semantics, to combine the different tasks of scene understanding, such as visual odometry (VO), video panoptic segmentation (VPS), \cite{yang2021tupper,yang2023unified,ying2023accurate,zhou2024hugs}, into a unified framework. While the VO and VPS performances of those methods need to be further improved to support real-world applications.

\IEEEpubidadjcol

Dynamic objects can not be negligible in the real-world applications. Some existing dynamic Simultaneous Localization and Mapping (SLAM) systems \cite{xiao2019dynamic,cui2019sof} propose to directly filter out potential dynamic objects, such as pedestrians or vehicles by semantic categories, but they overlook that stationary movable objects, like parked vehicles, also provide useful information for localization and tracking. Additionally, VPS frameworks \cite{kim2020video,qiao2021vip,ye2022hybrid,zhou2022slot,li2022video} focus on tracking individual instances in the dynamic scene without explicitly distinguishing whether objects are moving or stationary, but their tracking performance are still limited by the straight-forward modeling. In contrast, PVO \cite{ye2023pvo} proposes to integrate the VPS module with the VO module by combining optical flow and semantic segmentation results to predict the objects' movement, this integration realizes a more comprehensive scene modeling. However, the feature fusion of the PVO framework does not adequately consider the feature from adjacent frames and it directly adopts a rudimentary decoding network.

Furthermore, the sparse point clouds generated in urban scenarios result in a large number of holes and incomplete geometry shapes \cite{li2023read, yan2024street}, which is unable to provide the detailed and high-fidelity scenes that required by simulation environments of the autonomous driving \cite{zhou2024drivinggaussian}. Investigating the method to leverage Neural Radiance Fields (NeRFs) techniques for novel views synthesis of urban scenes is a promising direction to be explored.

\begin{figure*}[ht!]
    \centering
    \includegraphics[width=0.99\linewidth]{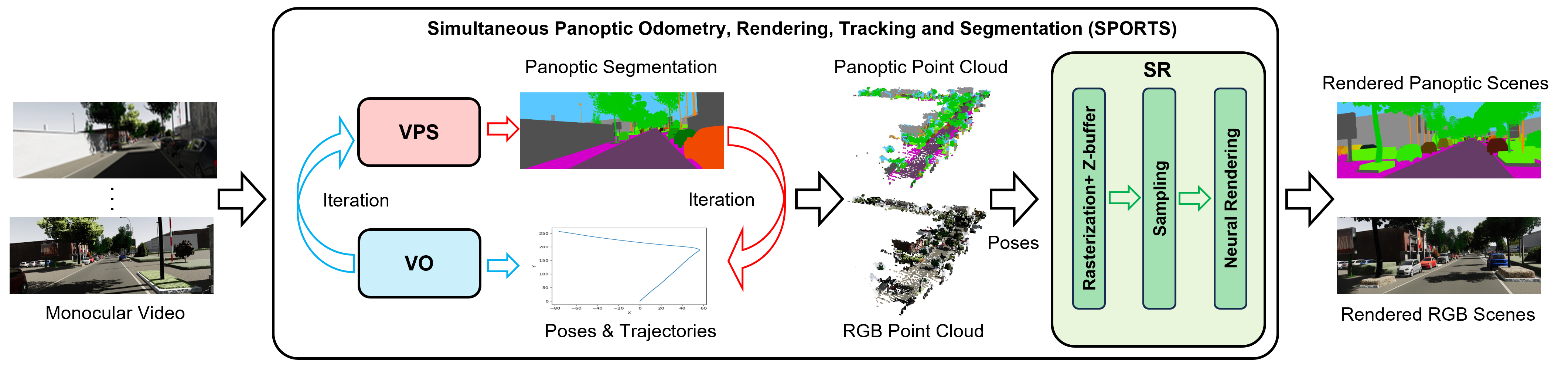}
    \caption{\textbf{The Overview of Our \FrameworkNM{} 
 Framework.} The framework solely receives monocular video as input, iteratively determines the temporal panoptic segmentation result, camera's poses and sparse point cloud maps, finally generates the photo-realistic and panoptic scene rendering result.}
    \label{fig:overview}
    \vspace{-0.5cm}
\end{figure*}

In this paper, we propose a novel framework, named \FrameworkNM{}, to achieve the comprehensive scene understanding, which tightly integrates video panoptic segmentation (VPS), visual odometry (VO), and scene rendering (SR) tasks through a unified perspective, the general diagram is shown in Fig. \ref{fig:overview}. In the VPS module, we employ a temporal kernel-based model as the prototype baseline, which can be replaced by other off-the-shelf video segmentation models.
Then, we introduce an attention-based adaptive geometry fusion mechanism that aligns multi-resolution features of the current frame with those of adjacent frames by utilizing poses, depths, and optical flows generated from VO. The aligning process can focus on key features in the fused feature maps, and automatically adjust the feature maps at different decoding stages. And we plug in a novel post-matching strategy for improving video instance tracking.
In VO, we obtain optical flow masks and dynamic masks by enrolling the panoptic segmentation results from VPS, which enhances the accuracy of pose estimation and the completeness of depth measurement. Lastly, the highly-precise pose estimation guarantee our point-based SR module to transform sparse point clouds into neural fields, which synthesizes the high-fidelity RGB images, and their corresponding panoptic-annotated views. 
Extensive experimental evaluations are conducted to validate the feasibility of this novel paradigm of holistic scene understanding and to demonstrate that \FrameworkNM{} framework outperforms other existing scene understanding methods.

Our contributions are summarized as:
\begin{itemize}

\item We propose the first framework, named \FrameworkNM{}, to unify visual odometry, rendering, objects tracking, and panoptic segmentation tasks to mutually understand and synthesize the urban street scenes. Our source code is publicly available\footnote{https://github.com/Fireworker20th/SPORTS}.
\item We propose an attention-based geometric fusion mechanism and a post-matching strategy for video panoptic segmentation, which improves the segmentation and tracking quality by 3.07 \% when compared to the counterparts.
\item We propose a two-stage panoptic-refined flow-aware depth propagation module to enhance learning-based visual odometry, which is beneficial to render
more high-fidelity views than other methods.
\item We exploit the foundation model to create more evaluation datasets than existing methods, which validates the generalization ability of SPORTS for diverse urban environments.

\end{itemize}

\section{Related Work}
\label{sec::relat}

\subsection{Video Panoptic Segmentation}

Video panoptic segmentation (VPS) requires generating instance tracking IDs within the video and integrating
instance segmentation with semantic segmentation to produce panoptic segmentation result. VPSNet \cite{kim2020video} pioneered this novel task and introduced Video Panoptic Quality (VPQ) as an evaluation metric, which utilizes task-specific prediction heads from instance segmentation, semantic segmentation, and tracking, jointly trained to obtain panoptic video results. 

VIP-DeepLab \cite{qiao2021vip} proposes a depth-aware VPS network by incorporating additional depth information, which is the first to combine geometric information with panoptic segmentation. HybridTracker \cite{ye2022hybrid} introduces optical flow to track instances based on feature space.
PVO \cite{ye2023pvo} combines geometric information with panoptic segmentation by introducing an adaptive geometric information fusion module. It enhances segmentation consistency by using optical flow predictions to match instances.
Slot-VPS \cite{zhou2022slot} is the first end-to-end framework that uses a unified representation called panoptic slot to encode all panoptic entities in a video. It retrieves and encodes coherent spatial-temporal object information into the panoptic slots through a video panoptic retriever. 
Video K-Net \cite{li2022video} learns to simultaneously segment and track “things” and “stuff” in a video through simple kernel-based appearance modeling and cross-temporal kernel interaction, Quasi-Dense \cite{pang2021quasi}, to extract tracking embeddings as tracking heads.
We observed that existing methods have not fully utilized multiple geometric information. 
The current VPS methods, when incorporating geometric information, merely use simple convolutions to reduce the dimensionality of the feature map and select relevant features from associated frames. They ignored the local cross-channel correlations.

In recent years, numerous studies \cite{wang2020eca} on visual tasks have introduced attention mechanisms, which have evolved into a crucial component of neural networks. In the field of image segmentation, many studies \cite{li2021ctnet, weng2021stage} utilized channel attention mechanisms \cite{hu2018squeeze, woo2018cbam, wang2020eca} to optimize networks and demonstrated their effectiveness. 
It is evident that channel attention mechanisms can effectively relate context and fuse feature information, making them a promising direction for VPS tasks that require cross-frame integration.

\subsection{Visual SLAM}

Visual SLAM is an active field in both industry and academia, especially in the past decade. 
Existing SLAM systems can be broadly categorized into two types: geometric-based methods \cite{engel2014lsd, engel2017direct, zhang2021geometry, yuan2021rgb} and learning-based methods, such as CNN-SLAM \cite{tateno2017cnn}, and Luo et al. 's approach \cite{luo2018real} which integrates predicted depth results with monocular SLAM, ESP-VO \cite{wang2018end} which predicts poses directly from video sequences in an end-to-end manner, and Wang et al. 's method  \cite{wang2019improving} uses semantic segmentation to estimate drivable areas to reduce VO drift.

Considering above supervised-learning methods have demonstrated promising performance, more and more unsupervised-learning methods are being explored for visual odometry (VO). TartanVO \cite{wang2021tartanvo} proposed the first learning-based visual odometry model, featuring a scale loss function and incorporating camera intrinsic parameters into the model, demonstrating reliable performance. DROID-SLAM \cite{teed2021droid} utilized the optical flow to define geometric residuals in its optimization procedures. It recursively updates camera poses and pixel-level depths through dense bundle adjustment layers. PVO \cite{ye2023pvo} introduced a panoptic update module in the camera pose estimation, to reduce the interference from dynamic objects.
Depth, optical flow, and semantic features are incorporated into unsupervised learning approaches for VO, while a more efficient and robust fusion strategy is open to be explored.

There are also several new trends for the visual SLAM frameworks, such as the semantic mapping with the holistic scene understanding \cite{yang2023unified,ying2023accurate}, and implicit mapping \cite{zhu2024nicer} for high-fidelity visual maps. TUPPer-map \cite{yang2021tupper} employs a temporal panoptic module for urban scene reconstruction. HUGS \cite{zhou2024hugs} conducts a joint optimization of geometry, appearance, semantics, and motion to achieve the holistic understanding of urban scenes. Xu et al. \cite{xu2024unified} further unify interactive, referring, and open-vocabulary semantic segmentation tasks with the panoptic segmentation task. Neural Radiance Fields (NeRFs) \cite{mildenhall2021nerf} have shown decent results in the high-fidelity rendering and the novel view synthesis. DVN-SLAM \cite{wu2024dvn} leverages an attention-based module to fuse implicit features from local discrete planes and the global continuous representation. And methods, such as SNI-SLAM \cite{zhu2024sni}, DDN-SLAM \cite{li2024ddn}, NIS-SLAM \cite{zhai2024nis}, propose to integrate neural implicit representations into the semantic mapping frameworks. Although those methods are able to generate the smooth surface reconstruction, they are prone to the localization precision, and are challenging to accommodate the visual odometry requirements, which hinders the deployment of those methods in the large-scale physical urban scenarios.

\subsection{NeRF-based Large Scene Reconstruction}
Traditional large-scale scenes reconstruction and view rendering rely on Structure-from-Motion (SfM) \cite{agarwal2011building, westoby2012structure, pollefeys2008detailed, schonberger2016structure} and Multi-View Stereo (MVS) \cite{furukawa2010towards, goesele2007multi, shen2013accurate}. Neural Radiance Fields (NeRFs) \cite{mildenhall2021nerf} has been naturally extended to the mapping tasks of large-scale scenarios for its photo-realistic synthesis results. 
S-NeRF \cite{xie2023s} improves the camera pose's modeling for learning better neural representations from street views. 
Block-NeRF \cite{tancik2022block}, NeRF-XL \cite{li2024nerf}, and Mega-NeRF \cite{turki2022mega} partitions the scene spatially and trains separate NeRFs for each partition. 
Methods like Switch-NeRF \cite{zhenxing2022switch}, SUDS \cite{turki2023suds}, and RoDus \cite{nguyen2024rodus} are designed to learn urban scene decomposition of static and dynamic elements. Swift-Mapping \cite{wu2024swift} proposes a sparse hybrid sampling strategy and hierarchical latent vectors to achieve the online neural implicit dense mapping for urban scenes. However, aforementioned methods learn the implicit representation of a scene solely from input images and corresponding poses, fail to utilize the geometry information in a comprehensive manner.

Point-based rendering \cite{li2023point, ruckert2022adop, li2024dgnr} incorporates geometric priors from sparse point cloud information, enabling a more accurate and faster scene synthesis. 
NPBG \cite{aliev2020neural} synthesizes high-quality indoor views from point clouds by learning neural descriptors to encode local geometry and appearance. NPBG++ \cite{rakhimov2022npbg++} utilizes multi-view observations to further predict the neural descriptors for each point.
READ \cite{li2023read} proposes a rendering network to learn neural descriptors from sparse point clouds, enabling synthesis of large-scale scenes. Therefore, the point-based neural rendering method is a promising candidate to facilitate our framework to achieve photo-realistic dense rendering in the real-time.
\section{Methodology}
\label{sec::methodology}

\subsection{System Overview}
As shown in Fig. \ref{fig:overview}, our \FrameworkNM{} framework is fed with image sequences $\mathbf{I}_t$ of a monocular video, simultaneously reconstructs a sparse point cloud map, estimates the camera poses $\xi_t$ relative to the world map, and synthesizes the novel scene, from the point cloud $\mathbf{P}$. \FrameworkNM{} consists of three modules: the video panoptic segmentation (VPS), the visual odometry (VO), and the scene rendering (SR). The VPS module fuses the geometric information obtained from VO to output the panoptic segmentation results. The VO module performs the estimation of the camera poses, the depth, and the optical flows, and estimation performance is enhanced by harvesting the semantic information from the VPS module. These two modules mutually facilitate each other through recursive interaction. The SR module aims to synthesize the photo-realistic scenes through the obtained sparse point cloud map. The motivation of \FrameworkNM{} lies in strengthening the correlation between VO and VPS tasks, leading to the fine-grained 3D scene understanding.

\begin{figure*}[ht]
    \centering
    \includegraphics[width=0.99\linewidth]{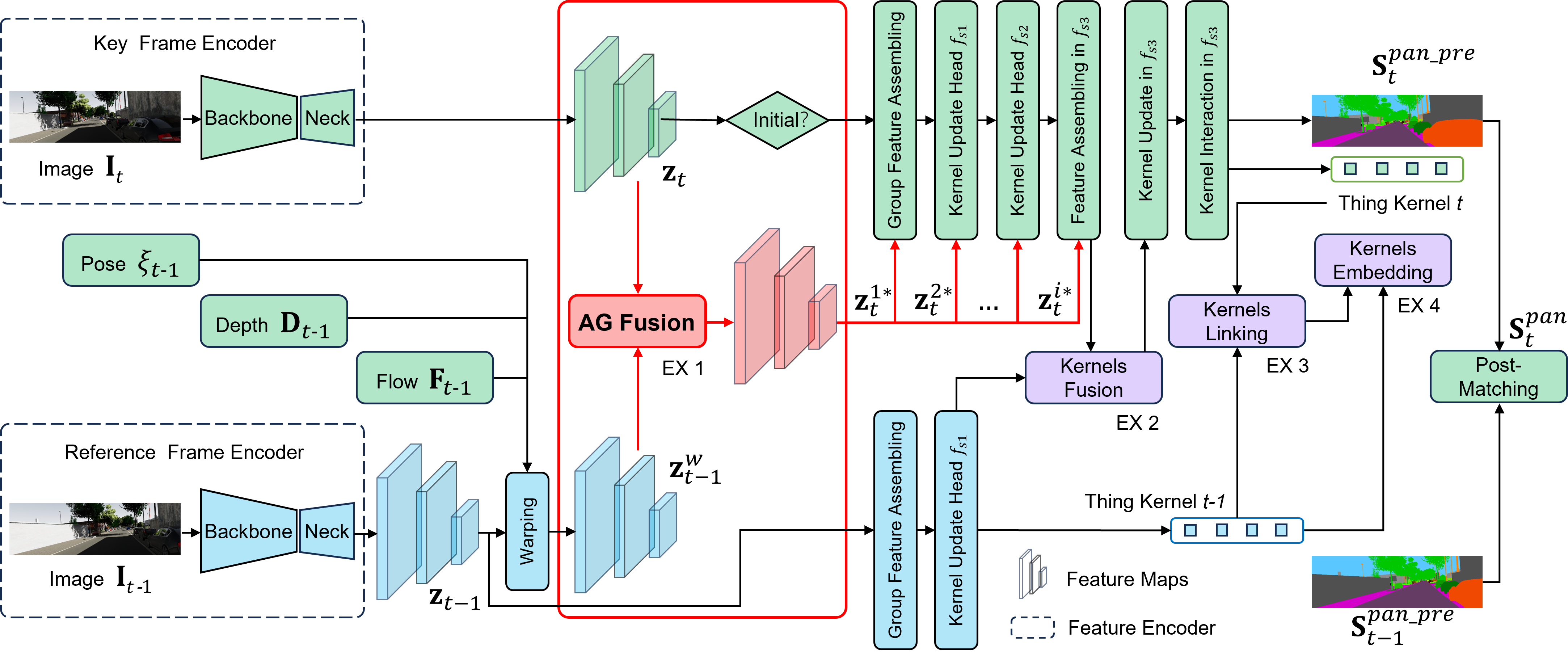}
    \caption{\textbf{The Diagram of Our Enhanced Video Panoptic Segmentation (VPS) Module.} We warp the feature map $\mathbf{z}_{t-1}$ generated by the neck of the reference frame $\mathbf{I}_{t-1}$ via leveraging poses, depth, and optical flow. The warped feature map $\mathbf{z}^w_{t-1}$ is then fused with the feature map $\mathbf{z}_{t}$ of the key frame $\mathbf{I}_{t}$ through $i$ fusion modules, resulting in i fused feature maps $\{\mathbf{z}^{1*}_{t}, \mathbf{z}^{2*}_{t}, ...,\mathbf{z}^{i*}_{t}\}$. This approach adaptively matches the feature maps with the kernel update heads by paying attention to the geometric knowledge, renamed as AG Fusion (Attention-based Adaptive Geometry Fusion, highlighted in red lines). Totally, our VPS module is enhanced via four levels of information exchanges (abbreviated as EX), including AG Fusion (EX 1), Kernels Fusion (EX 2), Kernels Linking (EX 3), and Kernels Embedding (EX 4), more details about AG Fusion is elaborated in Section \ref{sec::sec::AGFusion} and Fig. \ref{fig:FusionModule}.}
    \label{fig:VPSdiagram}
    \vspace{-0.5cm}
\end{figure*}

\subsection{Enhanced Video Panoptic Segmentation} 
The video panoptic segmentation processes monocular images input $\mathbf{I}_t$ to conduct instance segmentation, semantic segmentation, and instance tracking tasks, as shown in the Fig. \ref{fig:VPSdiagram}. The video panoptic segmentation result is also exploited for VO module in the next stage. 
Inspired by Video K-Net \cite{li2022video}, we leverage a group of learnable kernels to perform video panoptic segmentation. 

\subsubsection{\textbf{Image Panoptic Segmentation}}
Firstly, image-level panoptic segmentation step utilizes ResNet50, $f_{\theta_e}$, as the backbone network to extract multi-scale feature maps, $\mathbf{z}_t$, from images, which can be denoted as:
\begin{equation}
\mathbf{z}_t = f_{\theta_e}(\mathbf{I}_t)
\end{equation}

Then, a decoder $g_{\theta_d}$ with weights $\theta_d$ is used to generate the panoptic segmentation results, $S^{\text{pan}}_t$, which include both semantic segments $S^{\text{sem}}_t$ and instance segments $S^{\text{ins}}_t$. The panoptic segmentation result for each pixel $ \mathbf{u}$ is obtained via:
\begin{equation}
S^{\text{pan}}_t(\mathbf{u} \mid \mathbf{z}_t) = g_{\theta_d}(\mathbf{u},\mathbf{z}_t)
\end{equation}

The multi-scale feature maps $\mathbf{z}_t$ is fed into the fusion stage for tracking and odometry. And $S^{\text{pan}}_t$ is further utilized in the odometry module and scene rendering module. 

\subsubsection{\textbf{Kernels Embedding, Kernels Linking, and Kernels Fusion}} 
\label{sec::vps::keklkf} In order to enhance segmentation consistency between frames, kernels fusion and linking strategies inspired by \cite{li2022video} are utilized to obtain kernels embeddings. As shown in Fig. \ref{fig:VPSdiagram}, we first define a key image $\mathbf{I}^{\text{key}}_t$ from input images, reference images $\mathbf{I}^{\text{ref}}_t$ are randomly selected from its temporal neighborhoods, which is constrained by a window size $w$. The index of $\mathbf{I}^{\text{ref}}_t$ is within the range of $[t-w,t+w]$. 

Then, two kernels, $K_{\text{key}}$ and $K_{\text{ref}}$, are inferred correspondingly for feature embeddings, each kernel is responsible to predict a mask for either an instance or a stuff label. By following the Quasi-Dense \cite{pang2021quasi} learning, the matching of kernels between two sampled frames is deemed as positive if their corresponding masks have an IoU higher than $\alpha_1$, or negative if their IoU lower than $\alpha_2$. To be noticed, the IoU is calculated between the two regions that associated with the same object. And only the kernels matching the ground truth mask are optimized for mitigating the noise of the failure matching.

Assuming there are $V$ matched kernels, $\{K_v, v = 1, 2, ..., V\}$, as training samples on the $\mathbf{I}^{\text{key}}_t$, and $K$ matched kernels, $\{K_k, k = 1, 2, ..., K\}$, as comparison targets on the $\mathbf{I}^{\text{ref}}_t$, where $\{V < N \, | \, K_v \in K_{\text{key}}\}$, $\{K < N \, |\, K_k \in K_{\text{ref}}\}$, and $N$ is the total number of kernels. Then the loss function of our tracking task is summarized as:
\begin{equation}
\ \mathcal{L}_{\text{track}} = -\sum_{\mathbf{k^+}}\log  \frac{e^{\mathbf{v} \cdot \mathbf{k^+}}}{e^{\mathbf{v} \cdot \mathbf{k^+}} + \sum_{\mathbf{k^-}} e^{\mathbf{v} \cdot \mathbf{k^-}}},
\end{equation}

Here, $\mathbf{v}$, $\mathbf{k^+}$, $\mathbf{k^-}$ respectively denote the kernel embeddings of the training sample $K_v$, its positive target, and negative targets in $K_k$. 

An auxiliary loss is also designed below, where $c$ is 1 if the match of two samples is positive, and 0 otherwise.
\begin{equation}
\ \mathcal{L}_{\text{aux}} = {( \frac{\mathbf{v} \cdot \mathbf{k}}{||\mathbf{v}|| \cdot ||\mathbf{k}||} - c )^2},
\end{equation}

As shown in Fig. \ref{fig:VPSdiagram}, more information exchange of kernels are applied by following \cite{li2022video}, kernel linking is applied to perform interaction along the temporal dimension, and kernel fusion is leveraged to maintain segmentation consistency, and tackle the large motion and scale variation occurred in the scene. To be noticed, we employ the Video K-Net \cite{li2022video} model as the baseline of VPS, in the practical applications, other off-the-shelf segmentation models can be utilized by following the same fashion.

\subsubsection{\textbf{Attention-based Adaptive Geometry Fusion (AG Fusion)}}
\label{sec::sec::AGFusion}

\begin{figure*}[ht]
    \centering
    \includegraphics[width=0.99\linewidth]{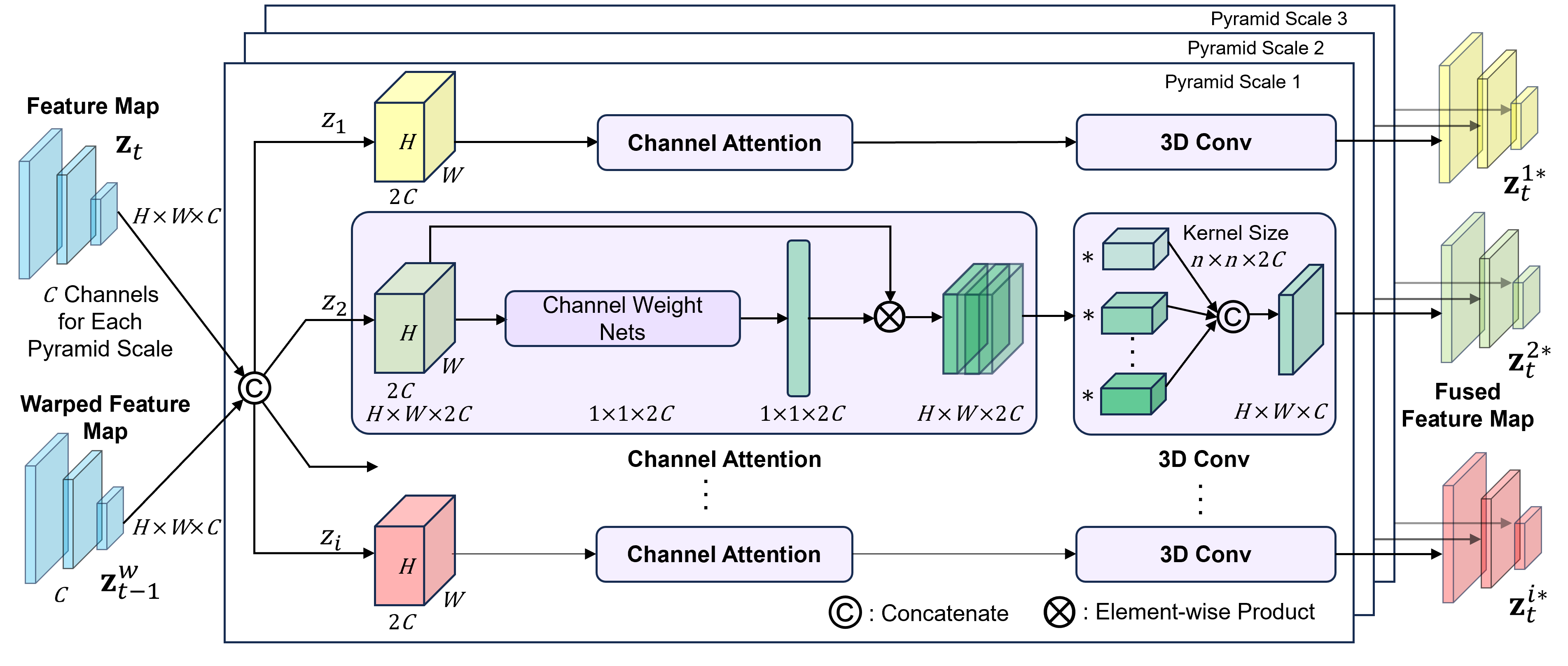}
    \caption{\textbf{An Illustration Diagram of Our Proposed AG Fusion Module.} $W$, $H$, and $C$ represent the width, height, and number of channels of the input features, respectively. The feature map $\mathbf{z}_t$ and the warped feature map $\mathbf{z}^\mathrm{w}_{t-1}$ are concatenated and made up by multiple pyramid scales. For each scale, the feature vector $\mathrm{z}_i$ is processed in parallel by $i_{th}$ branches, where the channel dimension of  $\mathrm{z}_i$ is 2$C$ as well. The $\mathrm{z}_i$ is re-weighted using channel-attention mechanism and then passed through a convolutional layer of kernel size $ n \times n \times 2C$. The fused feature map $\textbf{z}^{i*}_t $ for a branch is obtained by processing all the pyramid scale of feature maps.}
    \label{fig:FusionModule}
    \vspace{-0.5cm}
\end{figure*}

Video panoptic segmentation aims to obtain panoptic segmentation results $S^{\text{pan}}_t$ for each frame and maintain consistency across frames. FuseTrack \cite{kim2020video} attempts to enroll optical flow information for features fusion and track them based on feature similarity. PVO \cite{ye2023pvo} explores the depth information on top of camera poses and geometric information to enhance the perception of the 3D world.
However, these approaches overlook the correlation between cross-frame features and are unable to adjust for the network complexity of decoders, which hinders their performance in environments with complex scene changes.

The panoptic segmentation result $ S_{\text{pan}}^t $ is further utilized in both of the odometry and the scene rendering, while the multi-scale feature map $ z_t $ is fed into the fusion stage. During the fusion stage, depth $ D_t $, camera pose $\xi_t $, and optical flow $ F_t $ obtained from the Dense Bundle Adjustment Layer (DBA) in the VO module are used to warp the features from the previous frame to the current frame, generating a new fused feature map. The warping process is achieved through optical flow:

\begin{equation}
I_{t+1} (x) \approx I_t (x + f(x))
\end{equation}

Where the vector $ x $ represents the pixel position in the image. $ I_t (x) $ represents the pixel value of position $ x $ at $t_{\textbf{th}}$ frame, and $ I_{t+1} (x) $ represents the predicted pixel value at $t+1_{\textbf{th}}$ frame. The function $ f(x) $ denotes the optical flow field, describing the displacement vector from time $ t $ to time $ t+1 $.

With pose transformation, the depth-axis displacement of the same pixel in the next frame can be estimated, allowing for depth prediction in the subsequent frame. The warping order of each pixel in the optical flow field is determined based on depth, prioritizing pixels at larger distances. In the case of pixel warping conflicts, the later-warped pixel values overwrite the previous ones at the same coordinates, ensuring the accuracy of foreground details.

As shown in Fig. \ref{fig:VPSdiagram}, our attention-based adaptive geometry fusion module utilizes depth $\mathbf{D}_t$, pose $\xi_t$, and optical flow $\mathbf{F}_t$ information obtained from visual odometry (VO) module to map features from the previous frame $\mathbf{I}_{t-1}$ to the current frame $\mathbf{I}_t$, generates a warped feature map. By fusing the features of the current frame with the warped features, we obtain novel fused features for the current frame, called fused feature map. Here, the fused feature map is adaptively propagated to different stages of kernels updated module. 

The feature fusion procedures are further shown in Fig. \ref{fig:FusionModule}, $\mathbf{z}_t$ is the feature map at time $t$, and $\mathbf{z}^\mathrm{w}_{t-1}$ is the warped feature map at time $t-1$, we firstly warp the $\mathbf{z}_{t-1}$ at the pixel level using geometric information to obtain the warped $ \mathbf{z}_{t-1}^\mathrm{w} $, where $ \mathbf{z}_{t-1}^\mathrm{w} \in \mathbb{R} ^{H \times W \times C}$, C is the number of channels, H and W are height and width of feature maps. In the previous work like \cite{li2022video}, fused feature $\mathbf{z}^*_t$ is directly obtained through a convolutional layer with ReLU activation. However, this rudimentary method may struggle to handle complex scene changes. Therefore, we introduce a channel attention mechanism to perform weighted learning on the different channels of feature maps, enabling the fusion module to ignore unimportant features. 

We concatenate the warped $\mathbf{z}^\mathrm{w}_{t-1} $ with $ \mathbf{z}_t $ and learn channel weights through channel attention mechanism to obtain fused feature map $\mathbf{z}^*_{t}$:

\begin{equation}
\mathbf{z}^*_{t} = A \otimes (\circled{c} (\mathbf{z}_{t-1}^\mathrm{w}, \mathbf{z}_{t}))
\end{equation}

A is the computed attention weight, which is a vector of length 2 $\times$ C. Finally, a convolutional layer with kernel size $ n \times n \times 2C$ is used to obtain a fused feature map  $\textbf{z}^{i*}_t$ of size $ H \times W \times C$. The kernel size determines the receptive field of the convolutional network and increases with the enhancement of image resolution.  

VPS module employs multiple layers of Kernel Update Heads to continuously update kernels using feature maps. The key novelty is that we fuse the feature maps used by each layer of Kernel Update Heads separately, allowing the network to adaptively learn fusion strategies for different layers. And our channel weights are not dedicated to a specific channel attention network, it can be easily adapted to multiple Attention-based networks.

\subsubsection{\textbf{Loss Function and Training}}
Through the kernels embedding, the kernels linking, the kernels fusion, and the features fusion, our total loss function for enhanced video panoptic segmentation is represented as $\mathcal{L} = \lambda_{\text{cls}} \mathcal{L}_{\text{cls}} + \lambda_{\text{ce}} \mathcal{L}_{\text{ce}} + \lambda_{\text{dice}} \mathcal{L}_{\text{dice}} + \lambda_{\text{track}} \mathcal{L}_{\text{track}} + \lambda_{\text{aux}} \mathcal{L}_{\text{aux}},$ where  $\mathcal{L}_{\text{cls}}$ denotes the focal loss \cite{lin2017focal} designed for classification, $\mathcal{L}_{\text{ce}}$ and $\mathcal{L}_{\text{dice}}$ stand for the Cross-Entropy (CE) loss and Dice loss \cite{milletari2016v, wang2020solo} designed for segmentation. As described in Section \ref{sec::vps::keklkf}, losses $\mathcal{L}_{\text{track}}$ and $\mathcal{L}_{\text{aux}}$ are designed exclusively for instance trackings. The target assignment adopts the Hungarian assignment strategy \cite{carion2020end} by executing the end-to-end training paradigm.

\subsubsection{\textbf{Post-Matching}}
We notice that PVO employs a simple IoU-based matching approach, which exhibits strong performance in adjacent frame matching but leads to error propagation and performs poorly on long video sequences. This not only degrades segmentation quality but also affects the localization accuracy of the VO module due to incorrect optical flow pixel matching. In contrast, the learning-based method used in Video K-Net performs better on long sequences but suffers from misclassification in adjacent frames. Using IoU to enforce adjacent-frame consistency can further degrade long-sequence performance due to uncertain erroneous frames.

\begin{figure}[ht]
    \centering
    \includegraphics[width=0.99\linewidth]{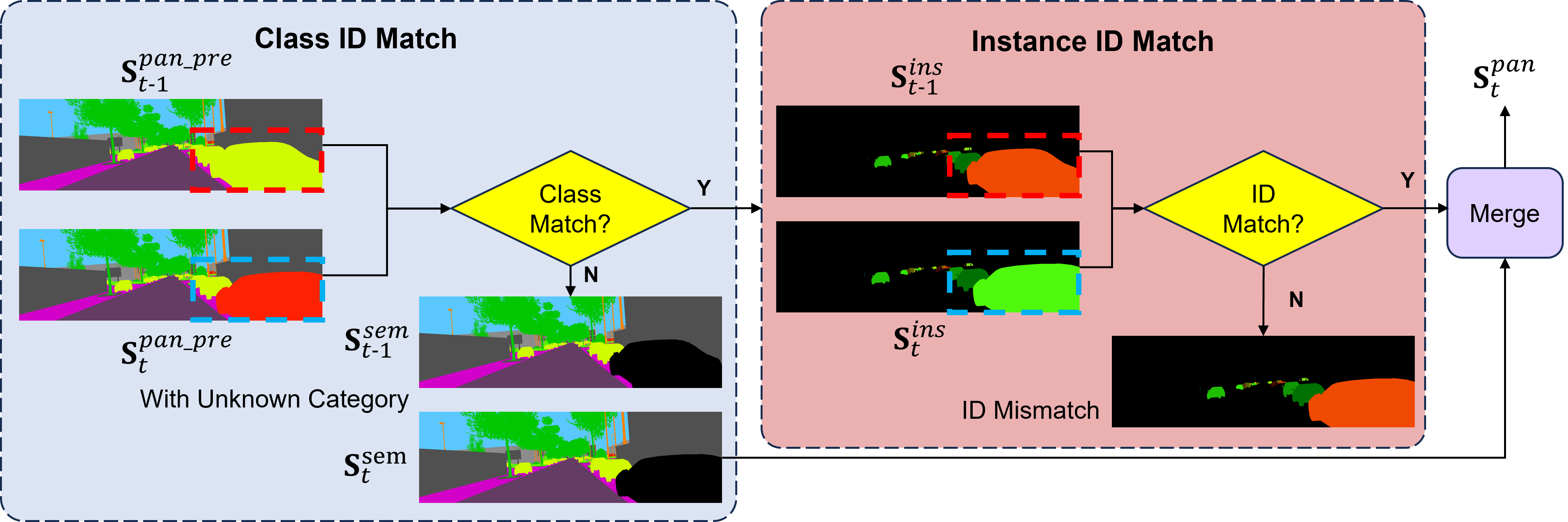}
    \caption{\textbf{An Illustrative Diagram of Our Post-Matching Process.}}
    \label{fig:match}
    \vspace{-0.5cm}
\end{figure}

To address these issues, we design a novel matching module to meet the requirements of both VPS and VO, as illustrated in Fig. \ref{fig:match}. First, we determine the semantic consistency of object categories between two frames based on IoU. If categories differ, we classify both as unknown to prevent further error propagation. If they match, checking their instance consistency, where instance IDs in the current frame are matched with those in the previous frame. Pixels classified as unknown are excluded from the optical flow matching process in VO as part of dynamic objects, thereby mitigating the impact of incorrect category assignments.

\subsection{Enhanced Visual Odometry}
\begin{figure*}[ht]
    \centering
    \includegraphics[width=0.99\linewidth]{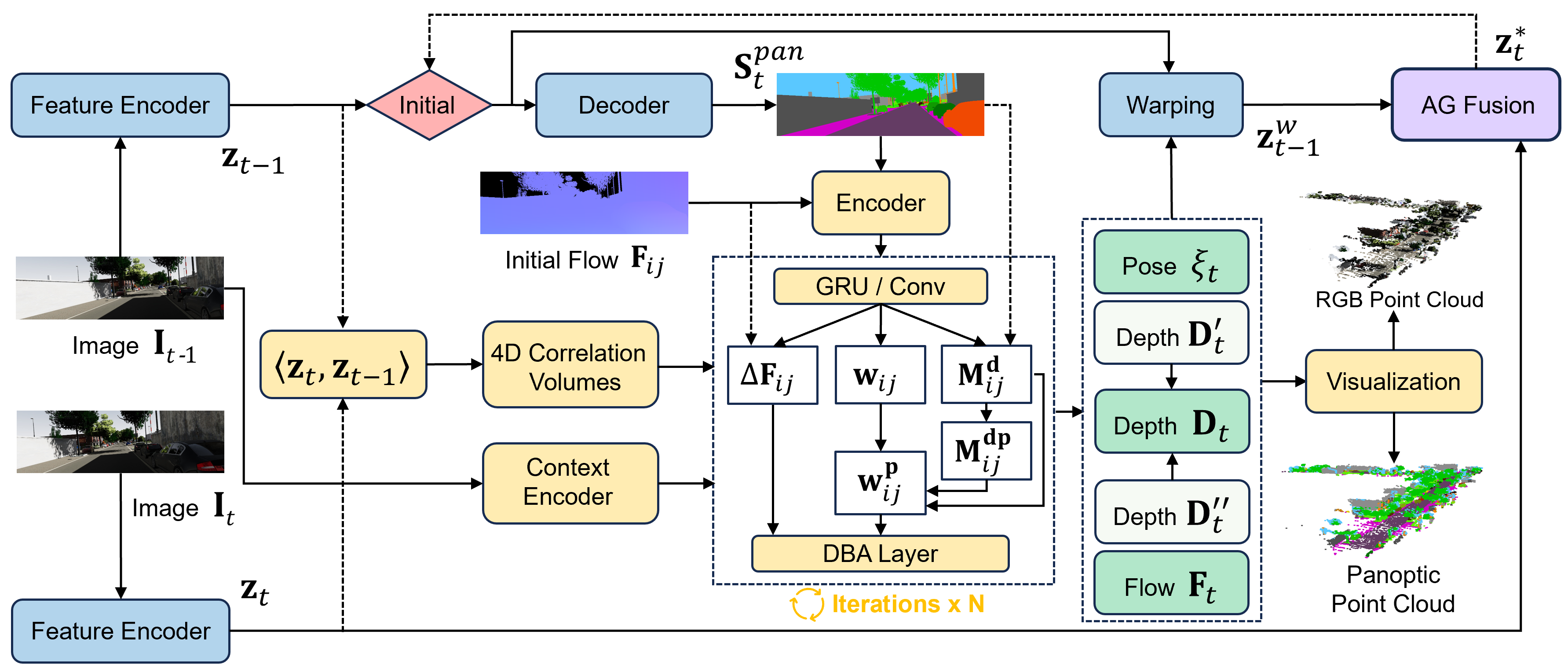}
    \caption{\textbf{The Diagram of Our Enhanced VO Module}. 
    Our method combines the video panoptic segmentation information, $S^{\text{pan}}_t$, with optical flow for the motion estimation, iteratively enhances the accuracy of camera pose estimation. The VO module outputs final pose, depth map and optical flow to facilitate the VPS in return.}
    \label{fig:VOdiagram}
    \vspace{-0.5cm}
\end{figure*}

In visual odometry, filtering out the interference of dynamic objects is crucial. The front end of DROID-SLAM \cite{teed2021droid} optimizes camera pose and depth residuals by iteratively optimizing optical flow residuals and confidence. Inspired by PVO \cite{ye2023pvo}, we also combine panoptic segmentation information, $S^{\text{pan}}_t$, with optical flow for the motion estimation, which helps to improve confidence estimation and thereby enhances the accuracy of camera pose estimation, the entire diagram is illustrated in Fig. \ref{fig:VOdiagram}.

The panoptic-aware flow features, the dual-frame 4D correlation volumes, and the features acquired by the context encoder are fed to the GRU and convolutional layers, generating a dynamic mask $ \mathbf{M}^{\mathbf{d}}_{ij} \in \mathbb{R}^{H \times W \times 2} $, a correlation confidence map $ \mathbf{w}_{ij} \in \mathbb{R}^{H \times W \times 2} $, and a dense optical flow delta $\Delta \mathbf{F}_{ij} \in \mathbb{R}^{H \times W \times 2} $. Panoptic-aware dynamic mask $\mathbf{M}^{\mathbf{dp}}_{ij}$ is obtained from $\mathbf{M}^{\mathbf{d}}_{ij}$ via fine-tuning the $S^{\text{pan}}_t$, stuff backgrounds are defined as static and foreground objects with higher dynamic probabilities are deemed as dynamic. 
Then, panoptic-aware confidence is defined as:
\begin{equation}
\mathbf{w}^{\mathbf{p}}_{ij} = \mathrm{\sigma}(\mathbf{w}_{ij} + (1 - \mathbf{M}^{\mathbf{dp}}_{ij}) \cdot \eta)
\label{equ::p-a-conf}
\end{equation}

$\eta$ is the scaling ratio of dynamic objects. The summation of the flow discrepancy $\Delta \mathbf{F}_{ij}$ and the raw optical flow $\mathbf{F}_{ij}$ is propagated to the Dense Bundle Adjustment (DBA) layer \cite{teed2021droid} to optimize the depth and poses. The foundation model is leveraged to estimate the training labels of depth maps, if the ground truth value of depth is not provided by the sensor.

The panoptic refinement procedure is iteratively executed via N times until the convergence. Following DROID-SLAM \cite{teed2021droid}, the residuals of the pose, the depth and the dynamic mask are respectively added to the current pose $\xi$, depth $\mathbf{D}$ and dynamic mask $\mathbf{M}^\mathbf{d}$:
\begin{equation}
 \mathbf{\xi}^{(k+1)} = e^{(\Delta\xi^{(k)})} \cdot \mathbf{\xi}^{(k)},
\end{equation} 

\begin{equation}
    \Theta^{(k+1)} = \Delta\Theta^{(k)} + \Theta^{(k)}, \Theta \in {\{\mathbf{D}, \mathbf{M^d}}\}.
\end{equation}
We concatenate $\mathbf{D}$ and $\mathbf{M}^d$ into $\Theta$. The correspondence is computed for each iteration using $\xi$ and $\mathbf{D}$. Given a pixel coordinate grid $ \mathbf{u}_i \in \mathbb{R}^{H \times W \times 2} $ in frame $ i $, we calculate its dense correspondence field $ \mathbf{u}_{ij} $ as follows:

\begin{equation}
    \mathbf{u}_{ij} = \Pi_{c}(\mathbf{\xi}_{ij}\cdot\Pi_{c}^{-1} (\mathbf{u}_{i}, \mathbf{D}_{i})),\mathbf{u}_{ij}\in\mathbb{R}^{H \times W \times 2},\\
\end{equation}
\begin{equation}
\mathbf{\xi}_{ij}=\mathbf{\xi}_{j}\cdot\mathbf{\xi}_{i}^{-1},
\end{equation}

For each edge $(i,j)\in \mathscr{E} $ in the frame graph as shown in DROID-SLAM \cite{teed2021droid}. $\Pi_{c}$ represents the camera projection, which maps a set of 3D points to the image, while $\Pi_{c}^{-1}$ refers to the re-projection which map the coordinate grid $ \mathbf{u}_i $ to the 3D point cloud. $\mathbf{u}_{ij} $ indicates the mapping procedure of pixel $ \mathbf{u}_i $ from frame $i$ to the coordinates of $ j_{th}$ frame.

The DBA layers map a set of latent vectors to a set of pose update $\xi^{'}$ and pixel depth update $\mathbf{D^{'}}$. The cost function of the whole frame is:

\begin{equation}
\mathbf{E}(\mathbf{{\xi}^{'}},\mathbf{D^{'}}) = \sum_{(i,j)\in \mathscr{E}} \Vert\mathbf{u}_{ij}^*-\Pi_{c}(\mathbf{\xi}_{ij}^{'}\circ\Pi_{c}^{-1}(\mathbf{u}_{i},\mathbf{D}_{i}^{'}))\Vert_{\sum_{ij}}^{2},
\label{equ::DBA}
\end{equation}

\begin{equation}
\sum_{ij} = \mathrm{diag}\mathbf{w}_{\mathbf{u}_{ij}}
\label{equ::cstf-diag}
\end{equation}

The Equation \ref{equ::DBA} and Equation \ref{equ::cstf-diag} is optimized by leveraging the Gauss-Newton optimizer.

After incorporating panoptic segmentation information, the excluded regions due to dynamic objects generates vacuum areas in the depth map and leads to the relatively sparser point cloud in 3D. For a given frame at time $t$, by following the same procedure as described in Equation \ref{equ::p-a-conf} to Equation \ref{equ::cstf-diag}, we feed a dynamic mask without segmentation refinement into the DBA layers to obtain a denser depth map $D^{''}_t$. The blank regions $\mathcal{B}$ in $D^{'}_t$ are then filled with the corresponding pixel values from $D^{''}_t$ to generate a more complete depth map. The final map after above depth propagation is calculated as:

\begin{equation}
D_{t} = \sum_{(u,v)\in \mathcal{B}} D^{'}_t + D^{''}_t
\end{equation}

\subsection{Scene Rendering}

Considering the rendering fidelity and computational resources, our scene rendering module is motivated by READ \cite{li2023read} method. The sparse point cloud $\mathbf{P} = \{p_1, p_2,..., p_n\}$ are obtained via VO module from input driving scene images $\mathbf{I}$ and known camera parameters. Neural descriptors $\mathbf{N}^d = \{n^d_1, n^d_2,..., n^d_n\}$ are computed from the $\mathbf{P}$. Each 3D point $p_i = \{x_i, y_i, z_i\}$ is associated with a neural descriptor vector $n^d_i$, to encode the local geometry and photometric characteristics. The SR module consists of three parts, which are rasterization, sampling, and the $\omega - \text{net}$ \cite{cho2021rethinking} based rendering.

\subsubsection{\textbf{Rasterization}}
Given the intrinsic and extrinsic parameters of the camera, we follow previous work \cite{li2023read} by using an 8-dimensional neural descriptor to represent RGB values observed from different viewpoints of the scene. In the rasterization phase, images with size $H \times W$ are captured by pinhole camera $C$, we construct a pyramid of rasterized raw image $\{S_k\}_{k=1}^{K}$ (K=4 in all our experiments), $S_{k}$ has the spatial size of $\frac{H}{2^{k}}\times\frac{W}{2^{k}}$, which is formed by assigning the neural descriptor of the point passing the depth test to each pixel. Subsequently, it is projected onto pixels under the camera's projection. Essentially, the neural descriptor $n^d_i$ encodes the local 3D scene content around $p_i$ into a feature vector. Neural descriptors of occluded points are screened out via Z-buffer scheme, and the holes generated by sparsity of point cloud $\mathbf{P}$ is filled via cubic projection \cite{li2023read}.

\subsubsection{\textbf{Sampling}} 
Sampling for point clouds is necessary for driving scenarios with data collected from thousands of meters. Due to the uneven distribution of point clouds in the different driving scenes, we utilize the Monte Carlo method \cite{shapiro2003monte} to sample a large amount of driving scene data, which is defined as $S_e^* \gets \text{argTop}_nQ(\mathbf{I}_e)$. For the image set $ S_e $ in training phase $ e $, the quality $ Q(\textbf{I}_e) $ of an image $\mathbf{I}_e $ is calculated using the perceptual loss \cite{johnson2016perceptual}. We select the worst-performing $ n $ samples from each phase as the training data to improve the learning efficiency. The patch sampling strategy similar to READ \cite{li2023read} is also applied to manipulate the image resolution for reducing memory usage.

\subsubsection{\textbf{$\omega-\text{net}$ based Rendering}}
Sparse point clouds often contain holes and outliers, which can degrade rendering quality. The purpose of the rendering network is to learn reliable neural descriptors $N^d_i$ to represent the scene. However, neural descriptors $N^d_i$ learned from point clouds still suffer from holes and surface bleeding. 
Following previous work \cite{li2023read, cho2021rethinking}, we exploit the $\omega - \text{net}$ to fuse multi-scale features. Our model fuse features at different scales and at the same scale to infer the missing points in sparse point clouds, which takes advantage of the complementary information.

\section{Experiment}
\label{sec::exp}
In this section, we comprehensively evaluate the different modules of our \FrameworkNM{} framework and thoroughly validate their generalization capability across various datasets. Specifically, for video panoptic segmentation task, we evaluate the segmentation performance on the VKITTI2 \cite{cabon2020virtual} and the VIPER \cite{richter2017playing} datasets. For visual odometry and scene rendering modules, we examine two datasets with dynamic scenes, VKITTI2 \cite{cabon2020virtual}  and dynamic KITTI \cite{geiger2013vision}, to assess the precision of estimated camera trajectories and the fidelity of rendered scenes. Additionally, we perform ablation studies to manifest the effectiveness of our framework design.

\subsection{Experimental Setup}
\subsubsection{\textbf{Datasets}} We evaluate the performance of our framework on three public datasets: the KITTI dataset \cite{geiger2013vision}, the VKITTI2 dataset \cite{cabon2020virtual}, and the VIPER dataset \cite{richter2017playing}. The KITTI \cite{geiger2013vision} is a dataset that captures real-world traffic scenes, covering a wide variety of static and dynamic objects across different environments, including highways, rural roads, and urban settings. The VKITTI2 \cite{cabon2020virtual} is a synthetic dataset which consists of 5 cloned sequences from the KITTI tracking benchmark, including RGB images, depth data, class segmentation, instance segmentation, camera poses, optical flow, and scene flow data for each sequence. The VIPER \cite{richter2017playing} is a high-quality video benchmark that offers instance and semantic segmentation labels, along with optical flow and camera poses ground truth. 

\subsubsection{\textbf{Evaluation Metrics}} The commonly-used metrics are adopted to evaluate the performance of our framework. For the performance of the visual odometry (VO) module, we employ the Root Mean Square Error (RMSE) of the absolute trajectory error as the evaluation metric. To evaluate the rendering quality, we leverage Peak Signal-to-Noise Ratio (PSNR), Structural Similarity Index Measure (SSIM), and perceptual loss (VGG loss) as the evaluation criterion. Besides, we incorporate a perceptual metric called Learned Perceptual Image Patch Similarity (LPIPS) into the rendering evaluation. Additionally, for evaluating the quality of panoptic video segmentation, we utilize the Video Panoptic Quality (VPQ) metric to measure the segmentation performance.

\subsubsection{\textbf{Implementation Details}}
This section details the implementation of our framework, with a focus on the training process. We implement our model using PyTorch and the MMDetection toolkit. The network training of our three main modules, the video panoptic segmentation (VPS) module, the visual odometry (VO) module, and the scene rendering (SR) module, are divided into four stages. First, our kernes-based video panoptic segmentation module is trained on datasets such as VKITTI2, where the frame size is 375 × 1242. We set AG fusion kernel size is $ 3 \times 3 $. We adopt the AdamW optimizer with an initial learning rate of 1e-4, a weight decay of 0.0001, and a batch size of 8 images. The window size $w$ of reference searching is defined as 2. The values of $\alpha_1$ and $\alpha_2$ in the kernel embedding stage are set to 0.7 and 0.3 respectively. Then, the training of the VO module follows the PVO. The scaling ratio of the dynamic object $\eta$ is set as 10. We trained this module on the VKITTI2 dataset for 80,000 steps using an Nvidia RTX-4090 GPU, which took about two days. Next, during the training of the AG fusion module, real depth, optical flow, and pose information are used as geometric priors to align the features. The backbone network of the pre-trained panoptic video segmentation module is kept frozen, and only the fusion module is trained. We use the AdamW optimizer with an initial learning rate of 1e-5 and a batch size of 8 images for optimization. Lastly, the training of the scene rendering module is motivated by READ, with a learning rate of 1e-4. The input of the scene rendering module is the sparse point cloud obtained from the VO module. All the trainings described above are conducted on a single Nvidia RTX-4090 GPU.

\begin{figure}[b!]\scriptsize
  \centering
  \setlength{\tabcolsep}{0.7pt}
  \begin{tabular}{cc}
     \makecell{\includegraphics[width=.49\linewidth]{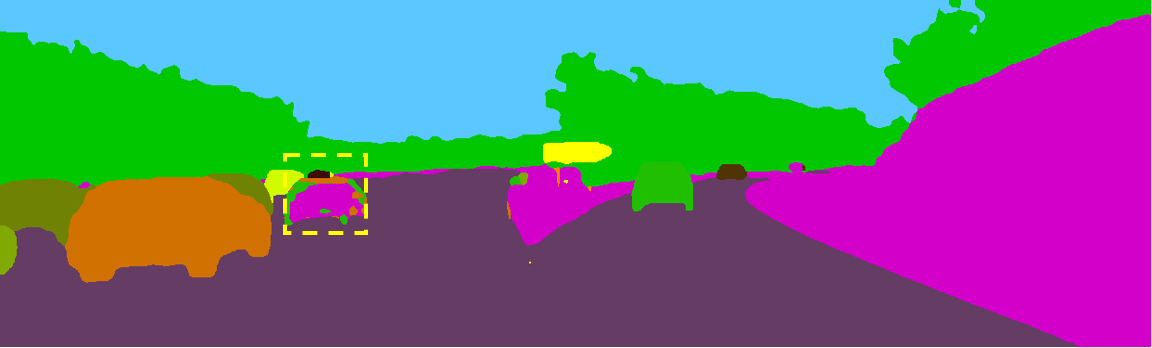}} & 
     \makecell{\includegraphics[width=.49\linewidth]{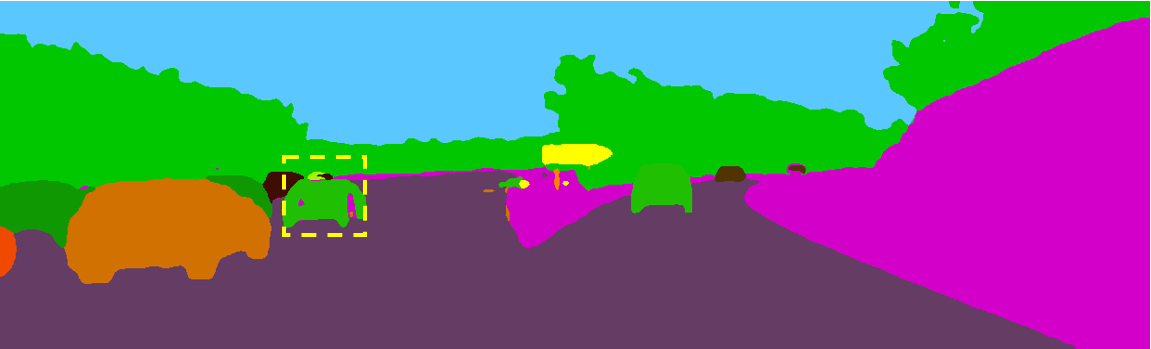}} \\
    \makecell{\includegraphics[width=.49\linewidth]{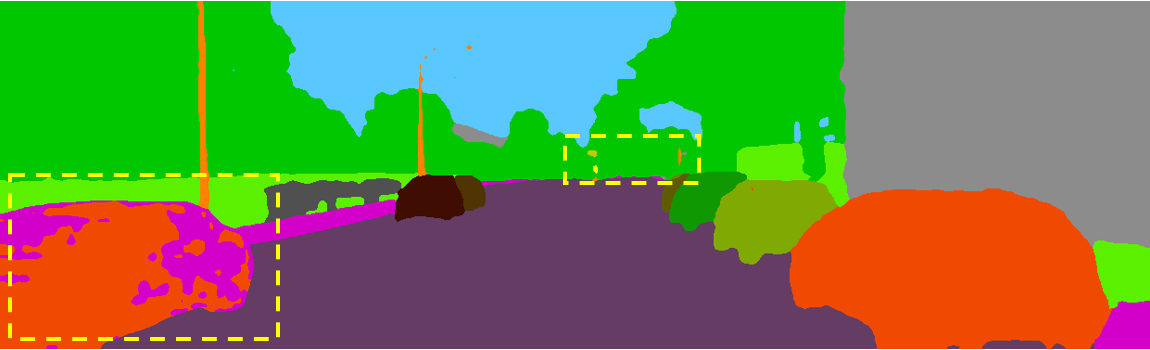}} &
    \makecell{\includegraphics[width=.49\linewidth]{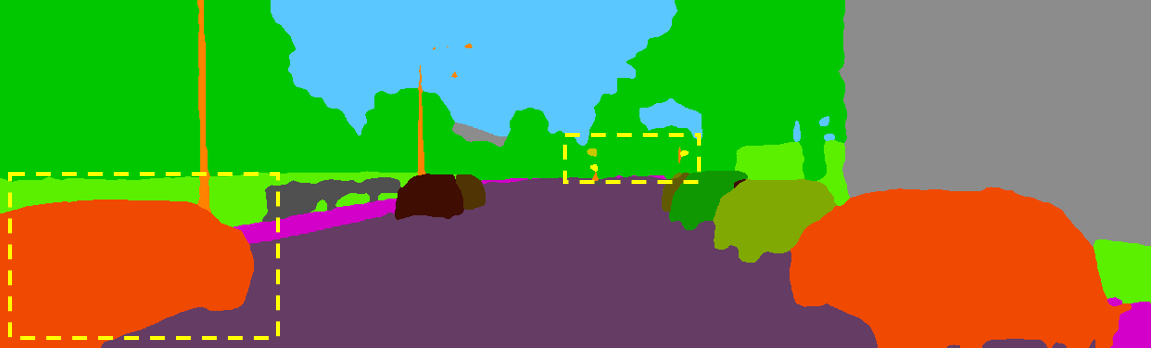}}\\ 
    \multicolumn{1}{c}{Video K-Net \cite{li2022video}} & \multicolumn{1}{c}{Ours} \\
  \end{tabular}
    \caption{\textbf{Visualization of segmentation improvements over the baseline method (Video K-Net).} The baseline method relied on the Video K-Net is prone to the classification failure and the inter-class fusion marked in yellow boxes. (\textbf{Best viewed in color and by zooming in.)}}
    \label{VPS baseline}
\end{figure}

\subsection{Results of the Video Panoptic Segmentation}
We select two instance-based video panoptic segmentation methods, PVO and Video K-Net, as baselines for the comparison. PVO integrates visual odometry and image panoptic segmentation tasks, enhances the cross-frame tracking performance with geometric information. Video K-Net learns to simultaneously segment and track \textit{things} and \textit{stuff} in a video using the kernel-based appearance modeling and the cross-temporal kernel interaction. Our focus is on comparing the adaptive geometry fusion (AG Fusion) module with other existing methods.

\begin{table*}[t!]
\centering
\caption{\textbf{The Quantitative Comparison of Video Panoptic Segmentation Performance on the VKITTI2 (val) Dataset.} For the PVO + AG Fusion method in the second row, we incorporate our adaptive geometry fusion mechanism to replace the fusion procedure of PVO. For the Video K-Net + PVO Fusion in the fourth row, we add the fusion mechanism on top of the original Vido K-Net architecture. And `Ours' stands for the proposed kernel-based method with an adaptive geometry fusion (AG Fusion) and post-matching strategy. Our method outperforms Video K-Net and PVO and their derived methods. `VK-Net' = `Video K-Net', `F.' = `Fusion', `Mem.' = Memory Usage, `Para.' = Parameter Number. Our frame rate is half of the PVO and VKNet methods with similar model size.}
\begin{tabular}{c| c| c| c| c| c | c  c  c}
\hline
\multirow{2}{*}{Methods} & \multicolumn{4}{c|}{Temporal Window Size} & \multirow{2}{*}{VPQ} & Mem. & Para. & FPS \\
\cline{2-5}

\multirow{2}{*}{} & k=0 & k=5 & k=10 & k=15 & & (GB) & (MB) &\\

\hline
PVO \cite{ye2023pvo} & 56.7 / 72.6 / 52.4 & 49.9 / 50.6 / 49.7 & 46.0 / 38.6 / 48.1 & 43.7 / 31.6 / 47.0 & 49.0 / 48.3 / 49.3 & 3.23 & 46.7 & 14.3\\

PVO \cite{ye2023pvo} + AG F.  & 57.4 / 72.7 / 53.2 & 51.5 / 54.6 / 50.6 & 47.6 / 42.5 / 49.0 & 45.4 / 35.3 / 48.1 & 50.5 / 51.3 / 50.2 & 3.23 & 45.79 & 14.3\\

VK-Net \cite{li2022video} & 60.7 / 58.4 / 61.3 & 56.2 / 47.6 / 58.5 & 53.7 / 40.1 / 57.4 & 52.3 / 35.3 / 56.9 & 55.7 / 45.3 / 58.4 & 3.99 & 41.55 & 13.3\\

VK-Net + PVO F.  & 60.3 / 57.0 / 61.3 & 55.8 / 45.3 / 58.6 & 53.3 / 37.9 / 57.5 & 52.1 / 33.6 / 57.1 & 55.3 / 43.4 / 58.6 & 4.18 & 42.73 & 7.2\\

\hline
Ours w/o Match  & 60.9 / 57.7 / 61.8 & 57.1 / 48.1 / 59.5 & 54.7 / 41.8 / 58.2 & 53.6 / 37.9 / 57.8 & 56.6 / 46.4 / 59.3  & 4.17 & 42.29 & 7.2\\

Ours & \textbf{61.5} / 57.2 / 62.6 & \textbf{57.5} / 46.6 / 60.5 & \textbf{55.1} / 40.2 / 59.1 & \textbf{53.9} / 36.3 / 58.7 & \textbf{57.0} / 45.1 / 60.2 & 4.19 & 42.29 & 7.1\\

\hline
\end{tabular}
\label{table_VPS_VKITTI2}
\end{table*}

\begin{table*}[t!]
\centering
\caption{\textbf{The Quantitative Comparison of Video Panoptic Segmentation Performance on the VIPER (val) Dataset.} The compared methods here keeps equivalent to Table \ref{table_VPS_VKITTI2}, it is obvious our proposed method achieves best performance on VIPER Dataset as well. The computational efficiency analysis for VIPER dataset is same to Table \ref{table_VPS_VKITTI2}. }
\begin{tabular}{c| c| c| c| c| c}
\hline
\multirow{2}{*}{Methods} & \multicolumn{4}{c|}{Temporal Window Size} & \multirow{2}{*}{VPQ}\\
\cline{2-5}

\multirow{2}{*}{} & k=0 & k=5 & k=10 & k=15 & \\

\hline
PVO \cite{ye2023pvo} & 23.4 / 10.3 / 33.5 & 21.8 / 7.1 / 33.1 & 20.7 / 5.1 / 32.7 & 20.2 / 3.7 / 32.8 & 21.5 / 6.5 / 33.0\\

PVO \cite{ye2023pvo} + AG Fusion  & 23.7 / 11.0 / 33.4 & 22.1 / 8.0 / 32.9 & 21.1 / 6.1 / 32.6 & 20.4 / 4.7 / 32.5 & 21.8 / 7.4 / 32.8\\

Video K-Net \cite{li2022video}  & 34.3 / 17.7 / 47.1 & 31.2 / 12.3 / 45.8 & 30.4 / 10.0 / 46.0 & 30.2 / 8.9 / 46.6 & 31.5 / 12.2 / 46.3\\

Video K-Net \cite{li2022video} + PVO Fusion  & 34.4 / 17.7 / 47.3 & 31.5 / 12.6 / 46.1 & 30.4 / 10.2 / 46.0 & 30.2 / 9.0 / 46.5 & 31.6 / 12.3 / 46.4\\

\hline
Ours w/o Post-Match  & 34.5 / 17.8 / 47.3 & 31.5 / 12.7 / 46.0 & 30.6 / 10.6 / 46.0 & 30.3 / 9.4 / 46.5 & 31.7 / 12.6 / 46.4\\

Ours  & \textbf{34.8} / 17.4 / 49.2 & \textbf{31.7} / 12.3 / 47.9 & \textbf{30.8} / 10.3 / 47.8 & \textbf{30.4} / 9.3 / 48.2 & \textbf{31.9} / 12.3 / 48.2\\

\hline

\end{tabular}
\label{table_VPS_VIPER}
\end{table*}

\begin{figure*}[t!]\scriptsize
  \centering
  \setlength{\tabcolsep}{0.7pt}
  \begin{tabular}{ccccc}
     \makecell{\includegraphics[width=.195\linewidth]{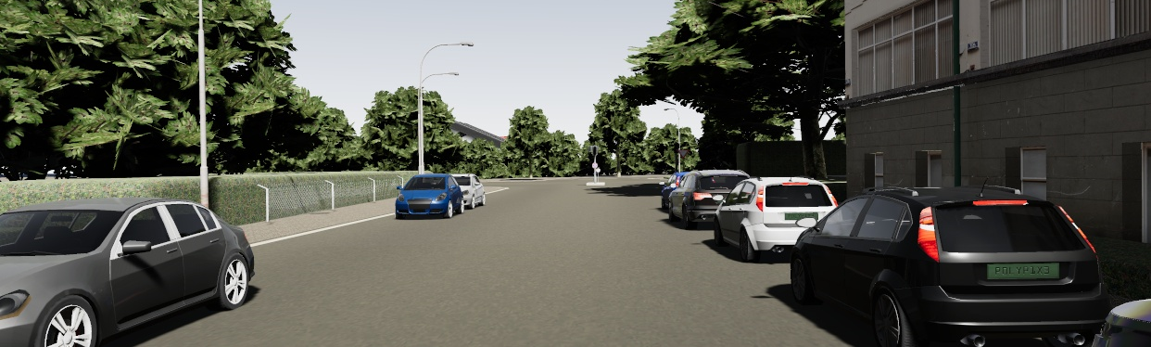}} & 
     \makecell{\includegraphics[width=.195\linewidth]{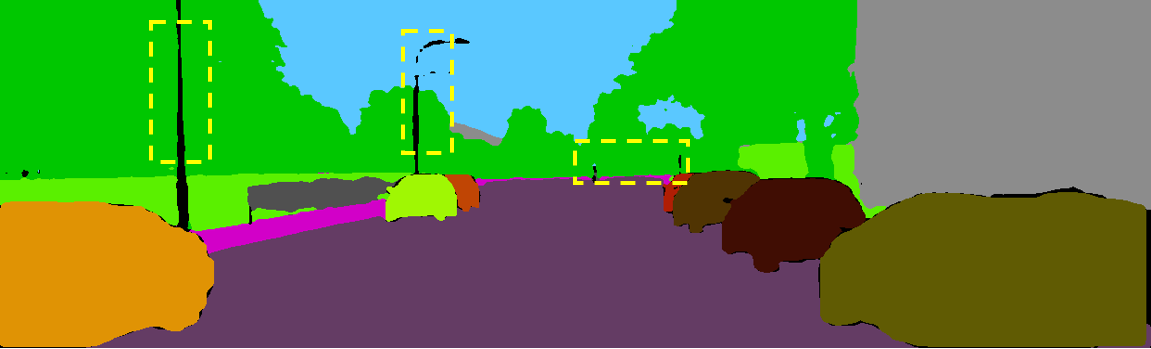}} & 
     \makecell{\includegraphics[width=.195\linewidth]{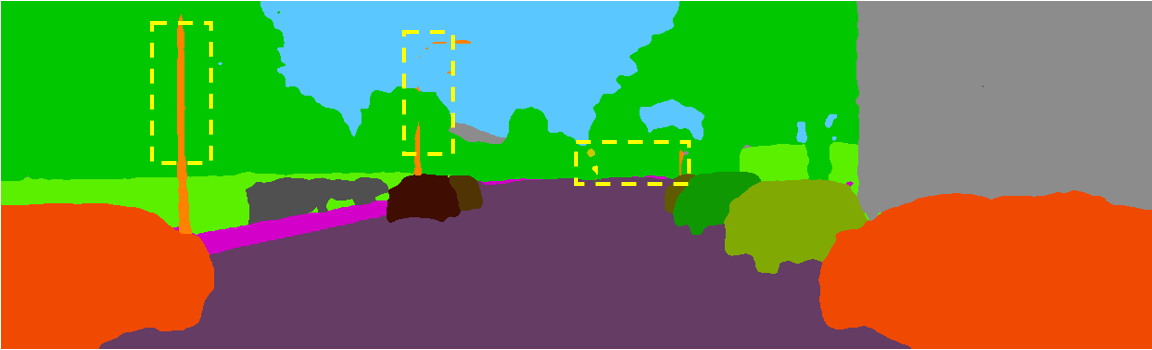}} & 
     \makecell{\includegraphics[width=.195\linewidth]{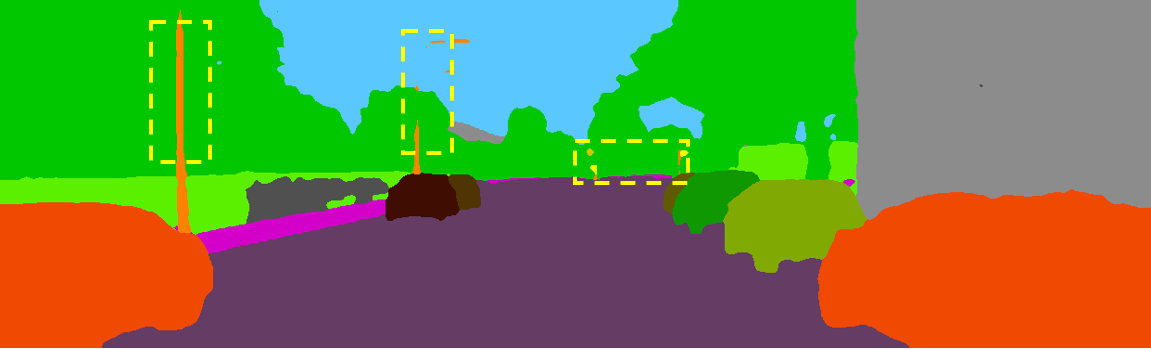}} & 
     \makecell{\includegraphics[width=.195\linewidth]{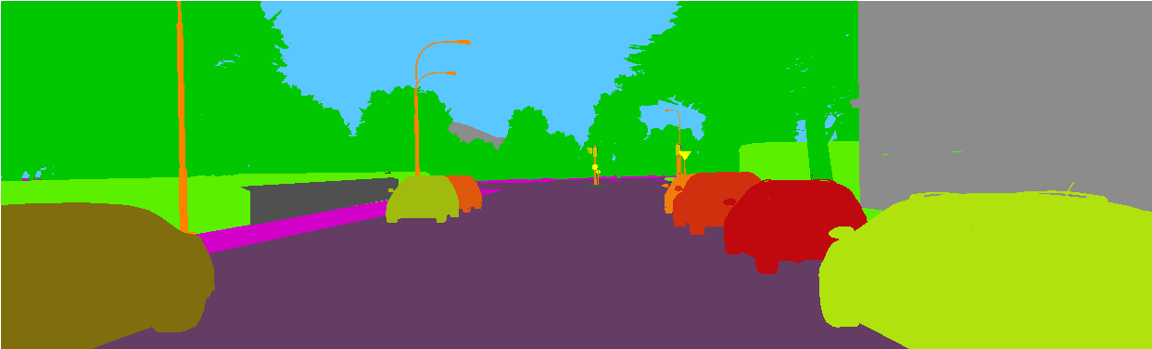}}\\
    \makecell{\includegraphics[width=.195\linewidth]{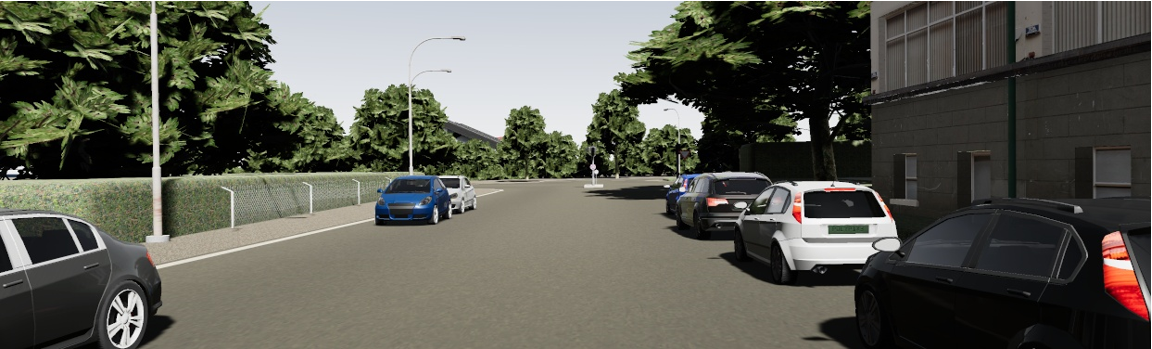}} & 
     \makecell{\includegraphics[width=.195\linewidth]{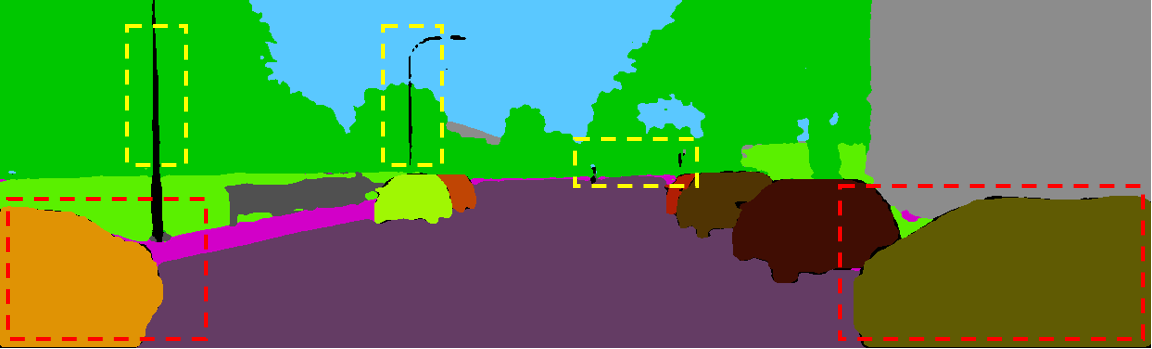}} & 
     \makecell{\includegraphics[width=.195\linewidth]{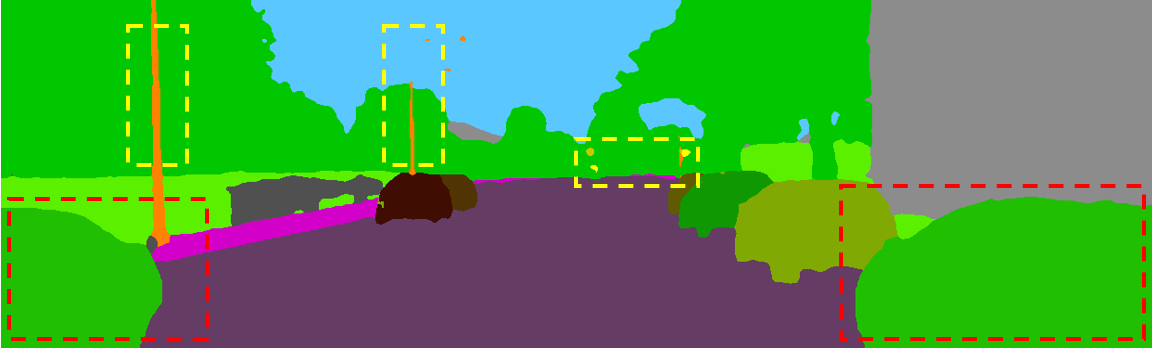}} & 
     \makecell{\includegraphics[width=.195\linewidth]{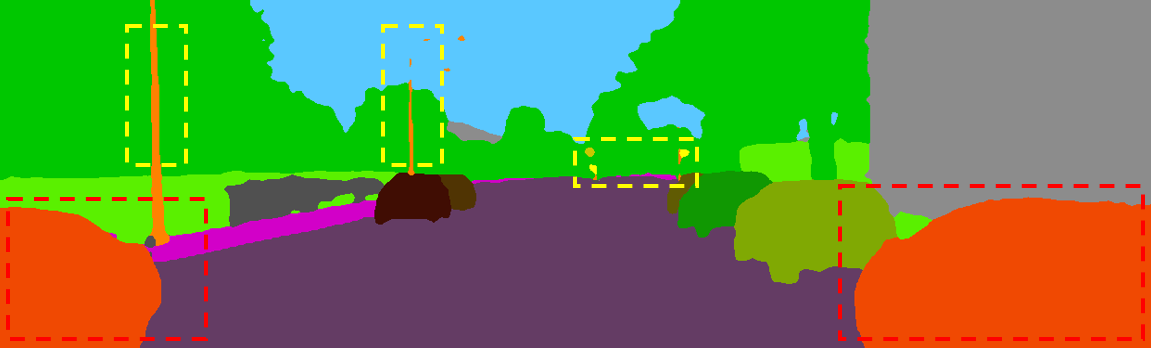}} & 
     \makecell{\includegraphics[width=.195\linewidth]{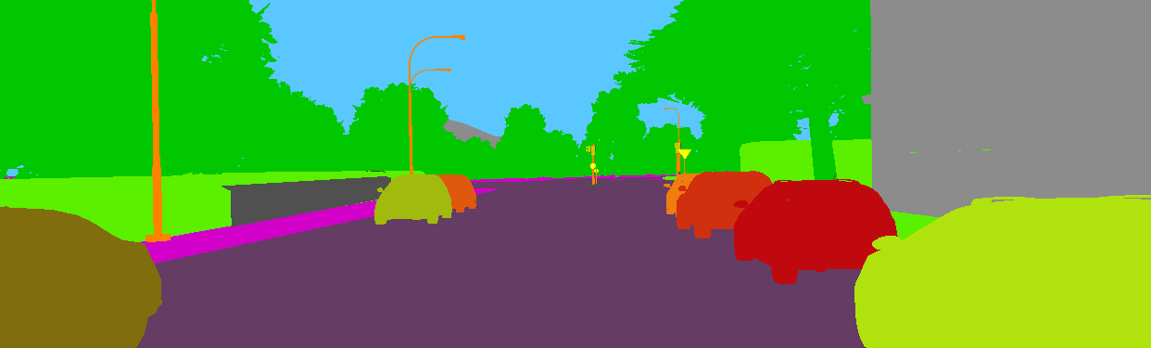}}\\
     \makecell{\includegraphics[width=.195\linewidth]{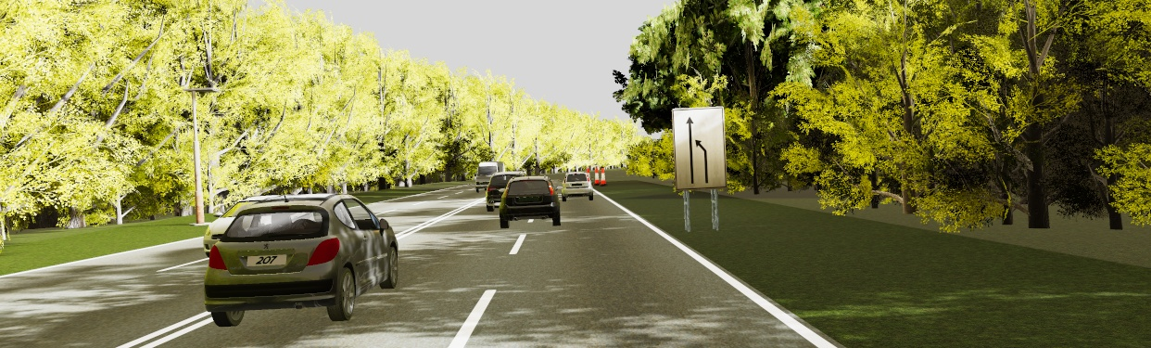}} & 
     \makecell{\includegraphics[width=.195\linewidth]{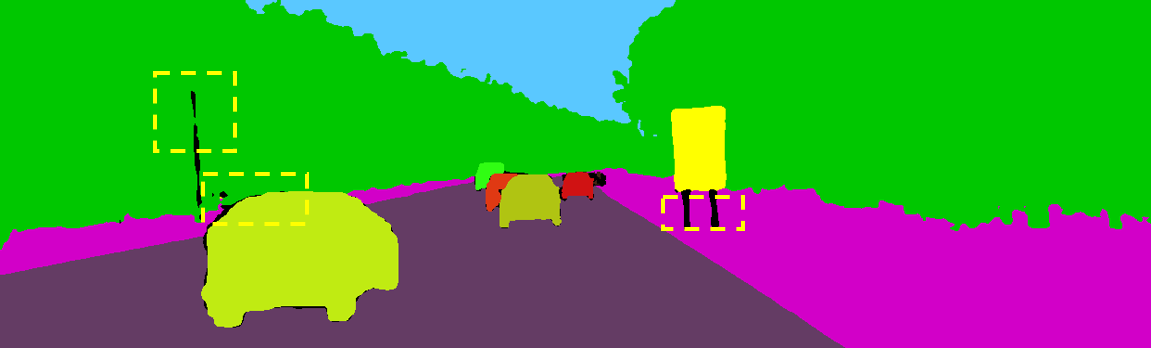}} & 
     \makecell{\includegraphics[width=.195\linewidth]{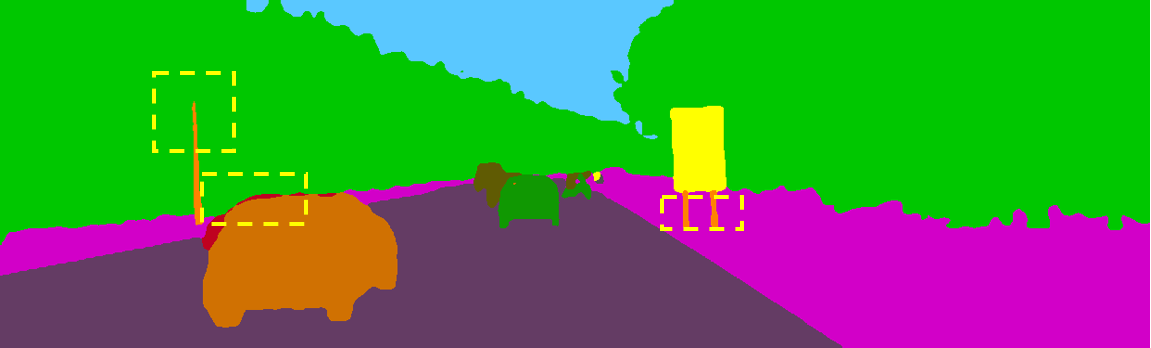}} & 
     \makecell{\includegraphics[width=.195\linewidth]{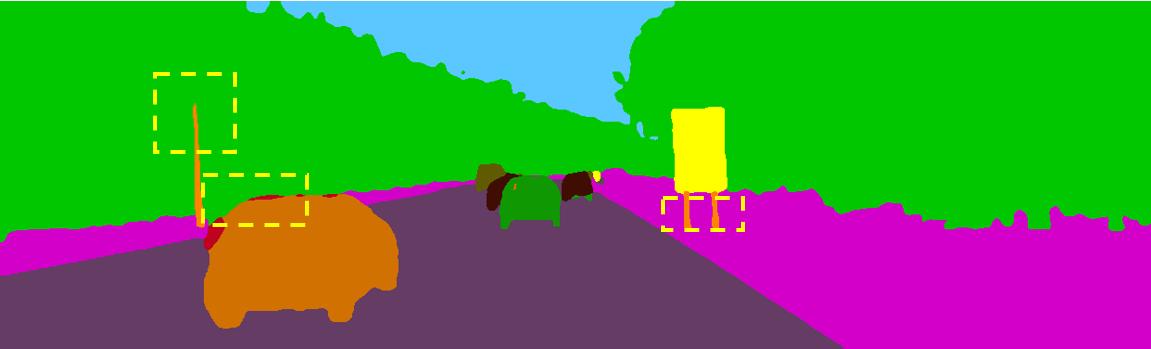}} & 
     \makecell{\includegraphics[width=.195\linewidth]{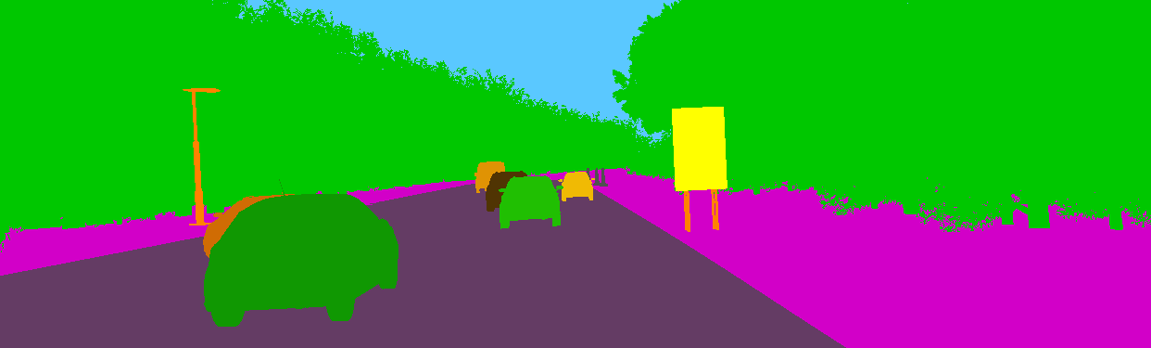}}\\
    \makecell{\includegraphics[width=.195\linewidth]{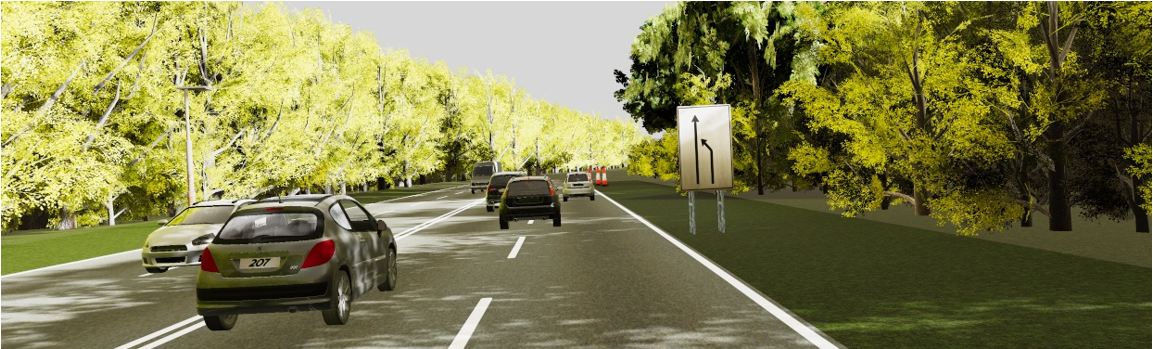}} & 
     \makecell{\includegraphics[width=.195\linewidth]{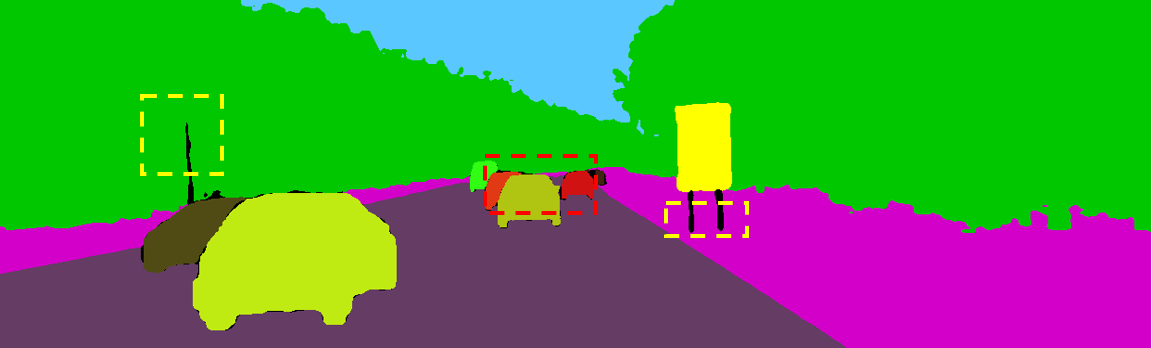}} & 
     \makecell{\includegraphics[width=.195\linewidth]{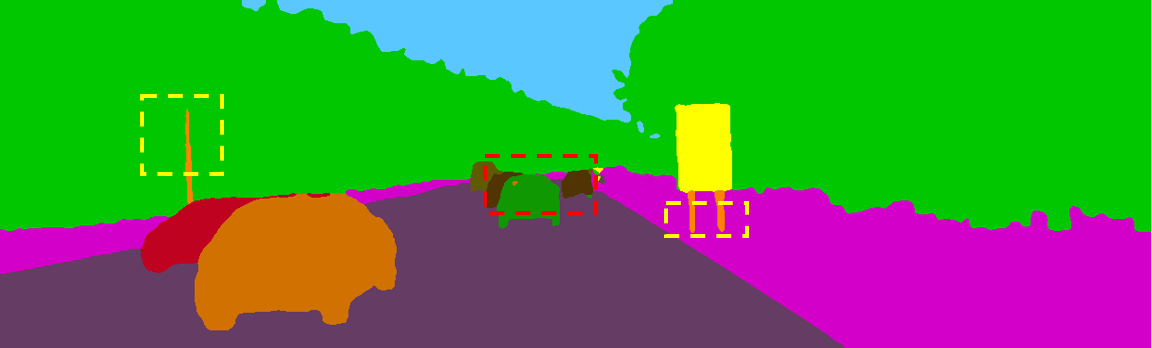}} & 
     \makecell{\includegraphics[width=.195\linewidth]{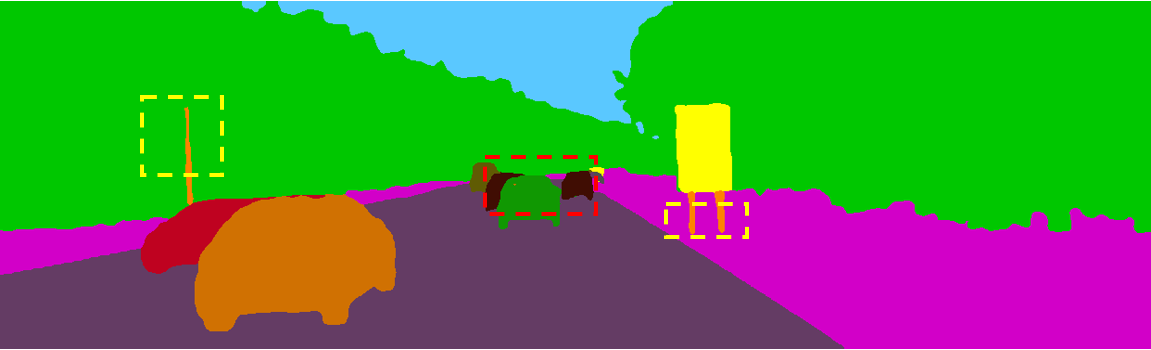}} & 
     \makecell{\includegraphics[width=.195\linewidth]{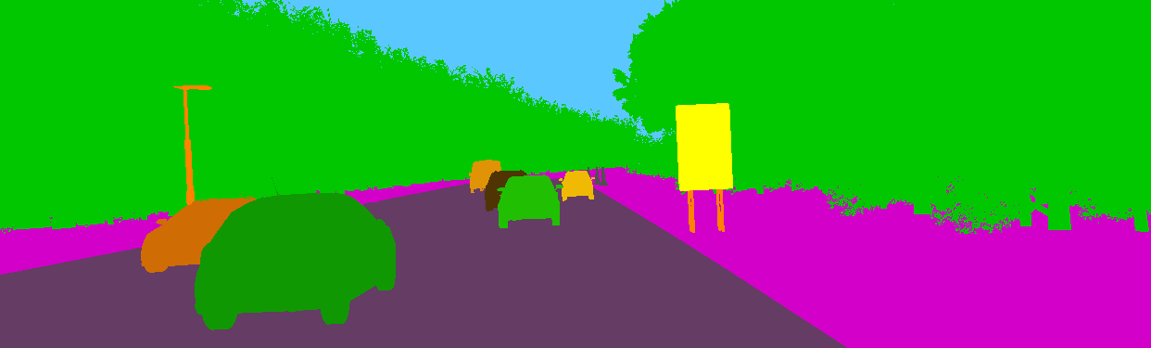}}\\
     \makecell{\includegraphics[width=.195\linewidth]{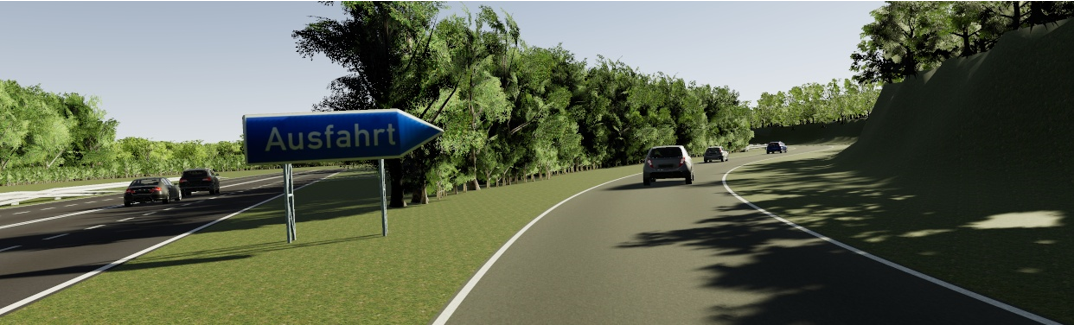}} & 
     \makecell{\includegraphics[width=.195\linewidth]{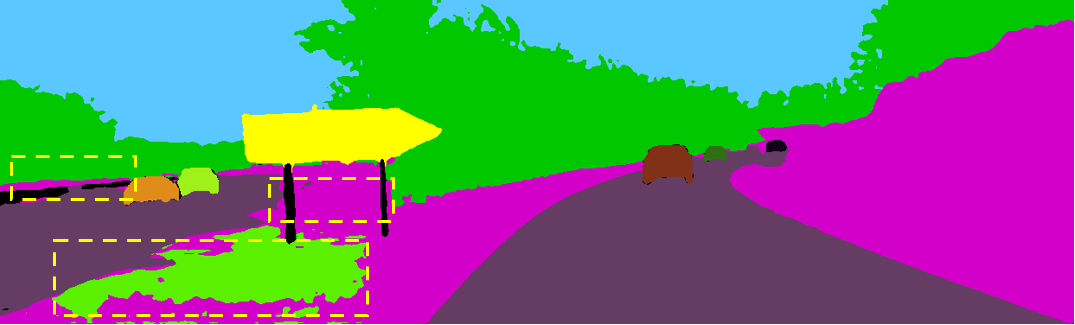}} & 
     \makecell{\includegraphics[width=.195\linewidth]{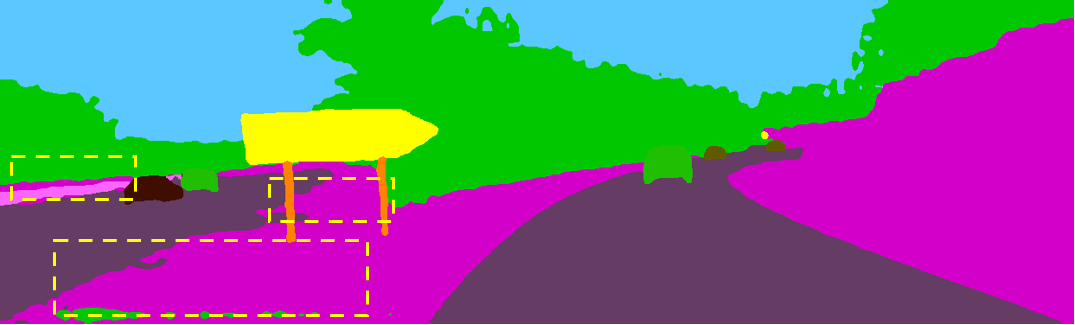}} & 
     \makecell{\includegraphics[width=.195\linewidth]{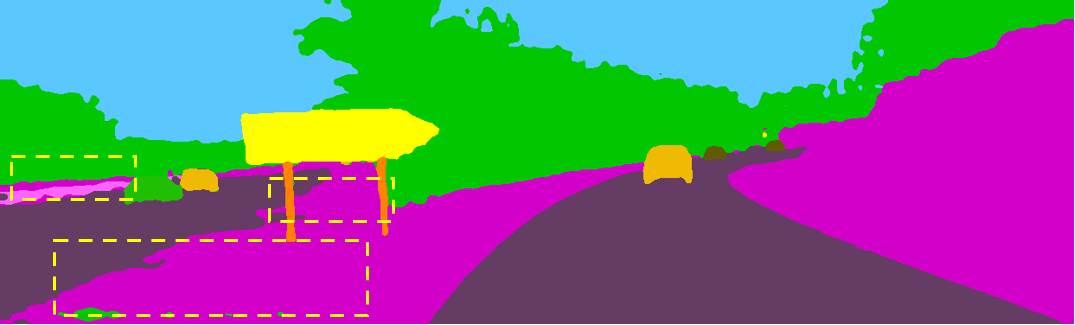}} & 
     \makecell{\includegraphics[width=.195\linewidth]{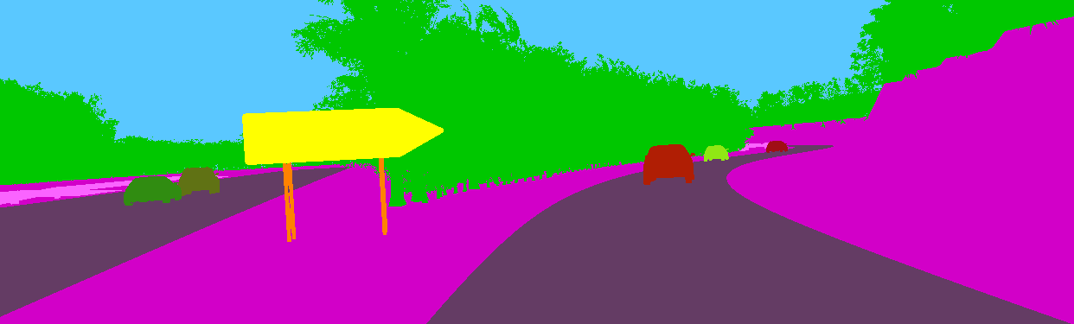}}\\
    \makecell{\includegraphics[width=.195\linewidth]{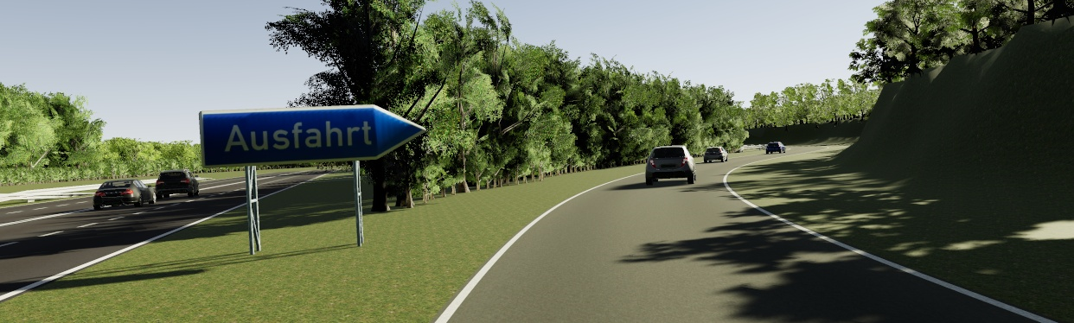}} & 
     \makecell{\includegraphics[width=.195\linewidth]{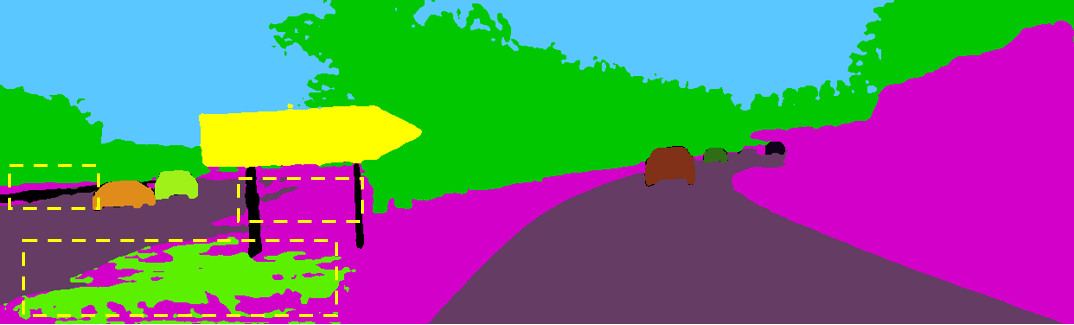}} & 
     \makecell{\includegraphics[width=.195\linewidth]{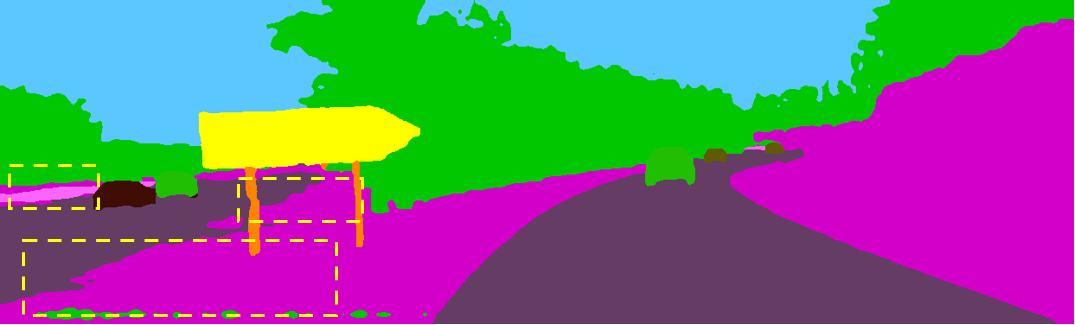}} & 
     \makecell{\includegraphics[width=.195\linewidth]{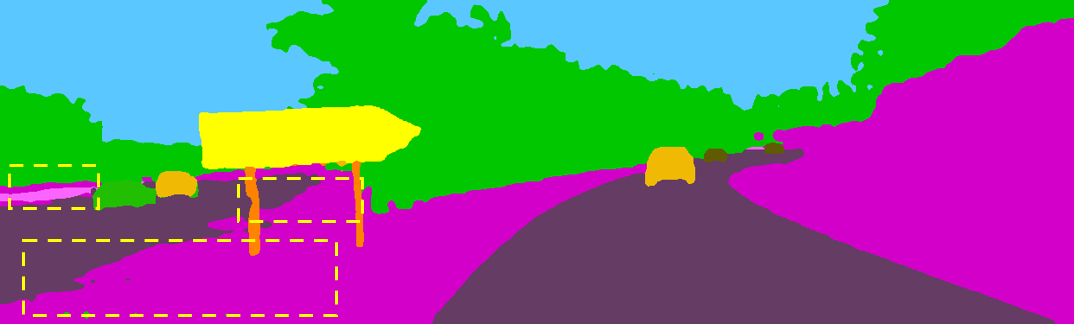}} & 
     \makecell{\includegraphics[width=.195\linewidth]{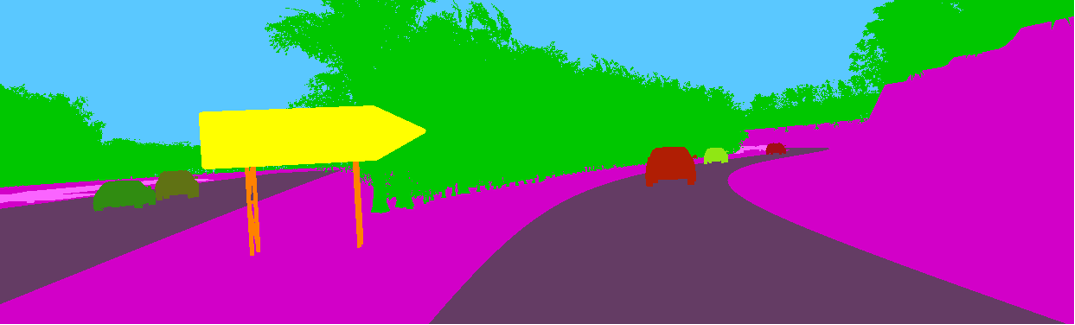}}\\
    \multicolumn{1}{c}{RGB image} & \multicolumn{1}{c}{PVO \cite{ye2023pvo}} & \multicolumn{1}{c}{Video K-Net \cite{li2022video}} & \multicolumn{1}{c}{Ours} & \multicolumn{1}{c}{Ground Truth}\\
  \end{tabular}
    \caption{\textbf{The Qualitative Results of Video Panoptic Segmentation on the VKITTI2 Seq 01 (The first two rows),  VKITTI2 Seq 18 (The 3rd and 4th rows), and VKITTI2 Seq 20 (The last two rows).} \textit{Stuff} backgrounds within same categories are painted with the same color, and \textit{thing} objects with the same predicted instance IDs are labeled with the same color. The box areas restricted by red dotted lines are corresponding to tracking errors for instance IDs, it manifests that Video K-Net sometimes failed to track vehicles on the boundary of the image (Red box area in the 2nd row), and both PVO and Video K-Net are struggling to track small vehicles (Red box area in the 4th row).
    And the box areas restricted by yellow dotted lines are classification failure (Street lamps and poles for PVO results) and inter-class confusion (Last two rows of PVO). (\textbf{Best viewed in color and by zooming in.)} }
    \label{fig:VPS_ALL}
\end{figure*}

\subsubsection{\textbf{VKITTI2 Dataset}} The synthetic KITTI2 dataset includes five cloned sequences from the KITTI tracking benchmark, providing RGB, depth, semantic segmentation, instance segmentation, camera poses, optical flow, and scene flow data for each sequence. Following PVO, we partition each sequence of the dataset into training, validation, and test sets at a ratio of 7:1:1. Our quantitative results are illustrated in Table \ref{table_VPS_VKITTI2}, our \FrameworkNM{} achieves the best performance in terms of VPQ quality. To be noticed, the fusion module of PVO exhibits a decrease in performance when combined with Video K-Net. However, our module not only outperforms PVO but also facilitates the original Video K-Net to further enhance the performance. And our post-matching strategy further refines the performance for instances registration. The VPQ value is increased by 3.061\% and 3.074\% respectively when combining our AG Fusion with PVO and Video K-Net, which brings the average boost of 3.07\%. As shown in Fig. \ref{VPS baseline}, the baseline methods using Video K-Net encounters inter-class confusion and classification failures, while our approach effectively obtains consistent classification results across frames. More comprehensive qualitative results for the temporal panoptic segmentation are shown in Fig. \ref{fig:VPS_ALL}, we compare our method with other two methods. Both PVO and Video K-Net exhibit the  classification failures, inter-class confusion, and instance tracking failures, while our method is able to segment more precise results when compared to the ground truth. To be noticed, the VPS results shown in Fig. \ref{fig:VPS_ALL} shows precision improvements for ID tracking of short video clips, while the long-term video panoptic segmentation via proposed method is still open to explore.


\subsubsection{\textbf{VIPER Dataset}} The VIPER is a high-quality video segmentation benchmark dataset that provides both instance segmentation and semantic segmentation labels. However, the geometric information of VIPER only includes ground-truth optical flow and camera pose data, without providing the depth information, and its optical flow data is irregularly discontinuous between frames. We reorganize the VIPER's labels and unify the instance and semantic labels into a panoptic segmentation label, and we utilize a foundation model called Depth-Anything \cite{yang2024depth} to estimate depth and deem it as the ground-truth data for training.

We select and pre-process the first six daytime scenes from VIPER for the training and the inferring. For the inferring, we utilize the last 30 frames from each video and we utilize 180 images in total. Due to the image resolution of VIPER is higher, we set the AG Fusion kernel size to \( 7 \times 7 \) to obtain the best performance, more details are discussed in section \ref{sec::ablation}. Table \ref{table_VPS_VIPER} summarizes the VPQ results by comparing with different methods on the VIPER dataset, where we observe a consistent result as shown in the VKITTI2 dataset. Due to the depth information of the VIPER is estimated by Depth Anything \cite{yang2024depth} algorithm, the performance on the VIPER manifest that our fusion strategy can be extended to the scenes without ground truth depth values.

\subsection{Results of the Visual Odometry}
In this section, the localization results on VKITTI2 and KITTI datasets are presented to validate the effectiveness of our enhanced VO.

\begin{table*}[t!]
\centering
\caption{\textbf{The VO Precision Comparison with Different Odometry Models on the VKITTI2 (val).} ATE metric is presented here, and the lower of ATE value is, the better of the odometry. From top row to the bottom row, the results of different methods are sorted from worst to the best based on the average RMSE out of the value of 5 sequences. (VPS$\rightarrow$VO) stands for the VPS executes one round of iterative optimization, (VPS$\rightarrow$VO $\times$ 2) stands for the VPS executes two rounds of iterative optimization, `Ours' stands for the combination between our kernel based feature extraction and our AG Fusion. Incorporating our AG Fusion mechanism boosts the performance of odometry, which can be observed via comparing PVO (VPS$\rightarrow$VO $\times$ 2) and PVO + AG Fusion (VPS$\rightarrow$VO $\times$ 2). mber. Our frame rate is slightly slower than PVO with the similar memory consumption.}
\begin{tabular}{c| c c c c c | c | c c}
\hline
Methods & vkitti01 & vkitti02 & vkitti06 & vkitti18 & vkitti20 & Avg & Memory Usage (GB) & FPS\\

\hline
DROID-SLAM \cite{teed2021droid} & 1.091 & \textbf{0.025} & 0.113 & 1.156 & 8.285 & 2.134 & 13.22 & 8.4\\

DOT\cite{ballester2021dot} & 1.140 & 0.140 & 0.070 & 1.000 & 9.120 & 2.294 & - & -\\

DytanVO \cite{shen2023dytanvo} & 10.654 & 5.262 & 1.226 & 8.397 & 18.290 & 8.766 & - & -\\

DynaSLAM\cite{bescos2018dynaslam} & 27.830 & - & - & - & \textbf{2.807} & -  & - & -\\

PVO \cite{ye2023pvo} (VPS\(\to\)VO) & 0.374 & 0.058 & 0.113 & 0.957 & 4.260 & 1.152 & 15.31 & 6.6\\

PVO \cite{ye2023pvo} (VPS\(\to\)VO $\times$ 2) & 0.372 & 0.057 & 0.113 & 0.952 & 3.808 &1.060 & 15.31 & 6.0\\
\hline
PVO \cite{ye2023pvo} + AG Fusion (VPS\(\to\)VO $\times$ 2) & 0.371 & 0.053 & 0.113 & 0.946 & 3.729 &1.042 & 15.95 & 5.8\\

Ours (VPS\(\to\)VO) & 0.369 & 0.055 & 0.113 & 0.951 & 3.290 & 0.956 & 15.95 & 5.4\\

Ours (VPS\(\to\)VO $\times$ 2) & \textbf{0.365} & \underline{0.051} & \textbf{0.113} & \textbf{0.943} & \underline{3.057} & \textbf{0.906} &15.95  &5.0\\

\hline

\end{tabular}
\label{table_VO_VKITTI}
\end{table*}

\begin{table}[h]
\centering
\caption{\textbf{The VO Performance Comparison with Different Odometry Models on the KITTI.} Similar to the experiment on VKITTI2 dataset, `Ours' stands for the combination between our kernel based feature extraction and our AG fusion. (VPS→VO × 2) is abbreviated to (× 2) for simplicity.}

\begin{tabular}{c| c c | c}
\hline
Method & KITTI09 & KITTI10 & Avg\\
\hline
ORB-SLAM3 \cite{campos2021orb}  & 64.74 & 80.17 & 72.455\\

DROID-SLAM \cite{teed2021droid} & 46.40 & 11.31 & 28.855\\

LIFT-SLAM\cite{bruno2021lift}  & 59.62 & 29.87 & 44.745\\

DynaSLAM \cite{bescos2018dynaslam}  & 41.91 & \underline{7.519} & 24.715\\

DSO \cite{engel2017direct}  & 28.1 & 24.0 & 26.050\\

DeFlowSLAM \cite{ye2022deflowslam} & 27.8 & \textbf{4.2} & 16.000\\

PVO \cite{ye2023pvo} (VPS\(\to\)VO) & 14.958 & 8.891 & 11.925\\

PVO \cite{ye2023pvo} ($\times$ 2) & 13.857 & 8.812 & 11.335\\
\hline
PVO \cite{ye2023pvo} + AG fusion ($\times$ 2) & \underline{13.432} & 8.592 & \underline{11.012}\\
\hline
Ours (VPS\(\to\)VO) & 14.438 & 8.706 & 11.572\\

Ours ($\times$ 2) & \textbf{12.793} & 8.572 & \textbf{10.683}\\

\hline
\end{tabular}
\label{table_VO_KITTI}
\end{table}

\subsubsection{\textbf{VKITTI2 Dataset}} We leverage the \textit{clone} sequence as the training set, the \textit{15-degree rotation} sequence as the validation set, and the \textit{30-degree rotation} sequence as the test set. As shown in Table \ref{table_VO_VKITTI}, we compare \FrameworkNM{} with four different VPS baseline models and their derived methods for the odometry. Our method achieves more convincing performance either with or without the iteration, it is obvious to observe that the odometry precision is increased with every one more iteration. Moreover, if our AG Fusion module is directly integrated into the segmentation module of the PVO \cite{ye2023pvo}, it also boost the odometry performance for PVO. And we also notice that our method outperforms the SLAM systems dedicated to the localization within dynamic scenes, DytanVO \cite{shen2023dytanvo} and DynaSLAM \cite{bescos2018dynaslam}. The memory usage and FPS results in Table \ref{table_VO_VKITTI} manifest the high demand of computation power for our VO method, we anticipate to accelerate it via patch-based optical flow matching in the future work.

\begin{figure}[b!]
    \centering
    \includegraphics[width=0.99\linewidth]{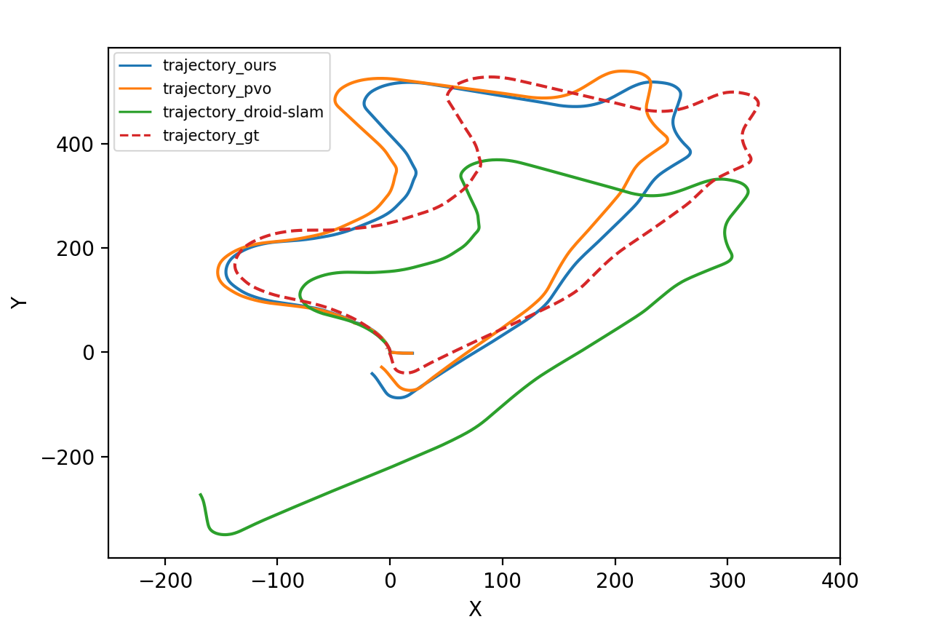}
    \caption{\textbf{Visualization of Estimated Trajectories of Three Methods on KITTI Sequence 09.} Different trajectories are aligned by same starting point (0,0). Our trajectory, drawn in the blue line, is the closest to the gt trajectory, drawn in the red dotted line.}
    \label{fig:VO_09}
\end{figure}

\begin{figure}[b!]
    \centering
    \includegraphics[width=0.99\linewidth]{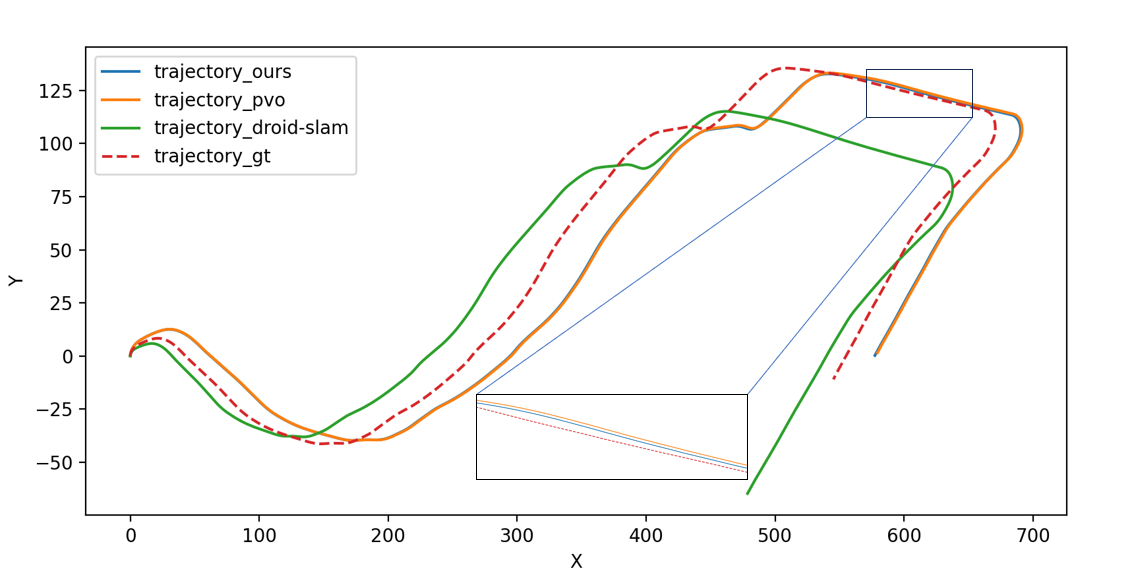}
    \caption{\textbf{Visualization of Estimated Trajectories of Three Methods on KITTI
Sequence 10.} The trajectory of sequence 10 is more smooth than sequence 09, by zooming into the detail, we can observe our \FrameworkNM{} generates the trajectory which is closer to the gt trajectory.}
    \label{fig:VO_10}
\end{figure}

\subsubsection{\textbf{KITTI Dataset}} We utilize sequences 09 and 10 from the KITTI dataset to validate the odometry accuracy of our system in more complex environments. We transfer the model trained on the VKITTI2 dataset to the KITTI dataset. As shown in Table \ref{table_VO_KITTI}, the pose estimation error of our method is further reduced, which is same to the VKITTI2 results. The trajectories comparison in Fig. \ref{fig:VO_09} and Fig. \ref{fig:VO_10} further visualize the advantage of our model. In the case of complex looped paths, such as sequence 09, our method generates the noticeably better performance. For simpler paths, like sequence 10, the trajectory difference needs to be zoomed in to observe, our approach maintains an advantage by a slight margin.

\subsection{Results of the Scene Rendering }
 
Our point-based rendering approach synthesizes sparse point cloud views into the denser presentation. The camera poses and point cloud are generated by the VO module of PVO and our \FrameworkNM{} respectively. 

\begin{table*}[t!]
\centering
\caption{\textbf{The Quantitative Comparison of Scene Rendering with PVO (Multiple Datasets).} Four aforementioned similarity metrics are evaluated with different test intervals. }
\label{table1_SR}

\begin{tabular}{c| c c c c| c c c c| c c c c}
\hline 
\multirow{2}{*}{} & \multicolumn{4}{c|}{KITTI 09} & \multicolumn{4}{c|}{KITTI 10} & \multicolumn{4}{c}{VKITTI2 20} \\
\multirow{2}{*}{} & VGG↓ & PSNR↑ & LPIPS↓ & SSIM↑ & VGG↓ 
& PSNR↑ & LPIPS↓ & SSIM↑ & VGG↓ & PSNR↑ & LPIPS↓ & SSIM↑\\
\hline
\multicolumn{13}{c}{Test every 100 frames}\\
\hline

PVO \cite{ye2023pvo} & 518 &22.64 &0.2549 &0.7150  & 551 &21.70 &0.2548 &0.6974 &519 &19.80 &0.2248 &0.7186 \\

Ours &\textbf{487} &\textbf{23.05} &\textbf{0.2301} &\textbf{0.7426} &\textbf{499} &\textbf{22.19} &0.2274 &\textbf{0.7436} &\textbf{485} &\textbf{20.09} &\textbf{0.2030} &\textbf{0.7381}\\
\hline
\multicolumn{13}{c}{Test every 10 frames}\\
\hline

PVO \cite{ye2023pvo} &517 &23.16 &0.2546 &0.7172 &547 &22.46 &0.2509 &0.6975 &510 &21.15 &0.2206 &0.6964 \\

Ours &\textbf{492} &\textbf{23.22} &\textbf{0.2333} &\textbf{0.7423} &\textbf{491} &\textbf{23.02} &\textbf{0.2193} &\textbf{0.7476} &\textbf{476} &\textbf{21.36} &\textbf{0.2032}  &\textbf{0.7269}\\

\hline

\end{tabular}
\end{table*}

\begin{figure*}[t!]\scriptsize
  \centering
  \setlength{\tabcolsep}{1pt}
  \resizebox{\textwidth}{!}{

  \begin{tabular}{cccccccccc}
      &\multicolumn{3}{c}{ KITTI Seq 09} & \multicolumn{3}{c}{ KITTI Seq 10} & \multicolumn{3}{c}{ VKITTI2 Seq 20}\\
      
      \makecell{\rotatebox{90}{PVO \cite{ye2023pvo}}}&{\makecell{\includegraphics[width=.30\linewidth]{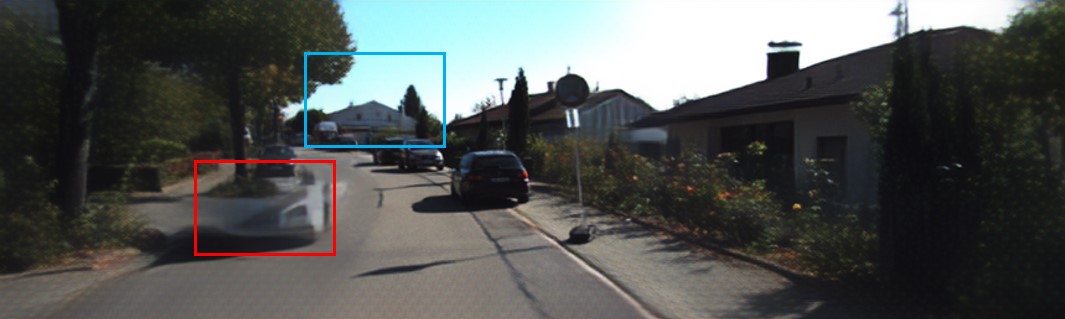}}} & 
     \multicolumn{3}{c}{\makecell{\includegraphics[width=.30\linewidth]{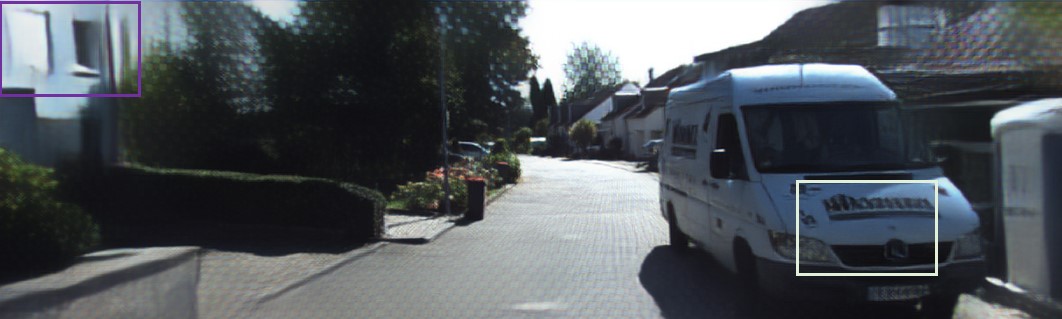}}} & 
     \multicolumn{3}{c}{\makecell{\includegraphics[width=.30\linewidth]{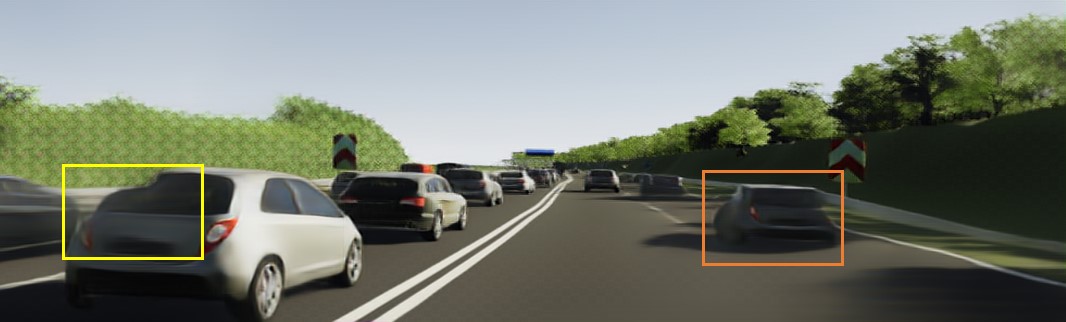}}}\\
    \makecell{\rotatebox{90}{Ours}}&
    \multicolumn{3}{c}{\makecell{\includegraphics[width=.30\linewidth]{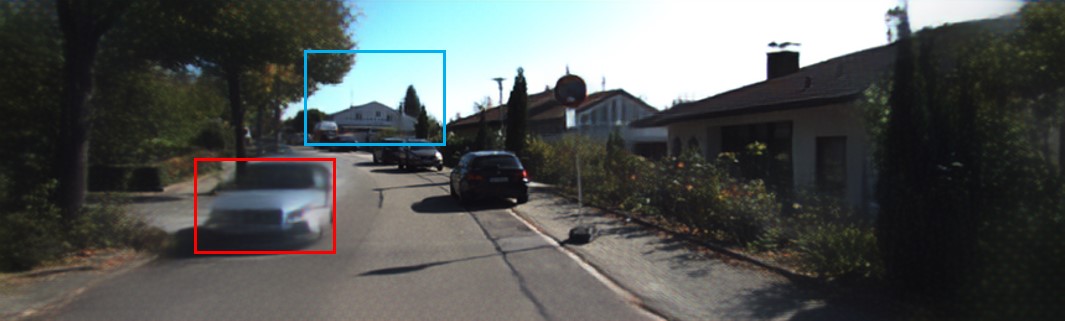}}} & 
     \multicolumn{3}{c}{\makecell{\includegraphics[width=.30\linewidth]{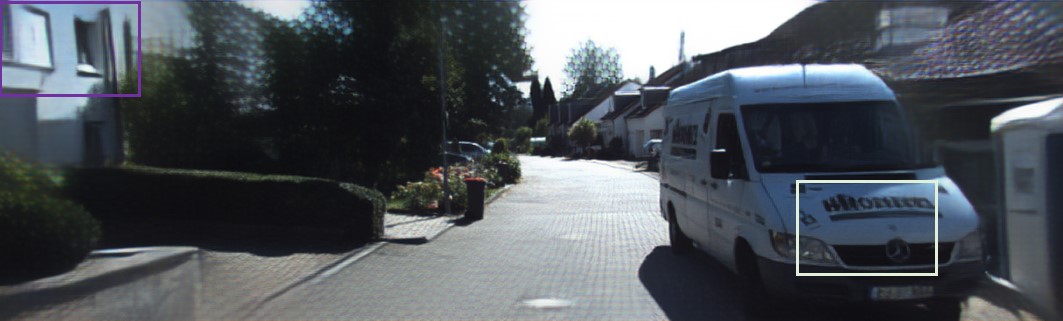}}} & 
     \multicolumn{3}{c}{\makecell{\includegraphics[width=.30\linewidth]{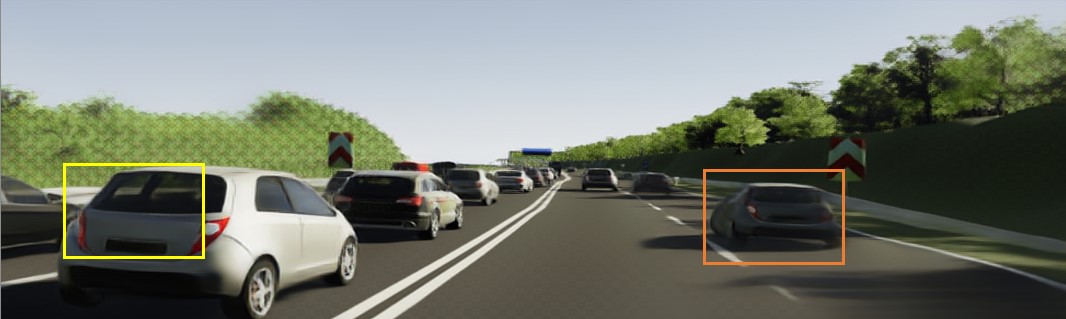}}}\\
     \makecell{\rotatebox{90}{GT}}&
     \multicolumn{3}{c}{\makecell{\includegraphics[width=.30\linewidth]{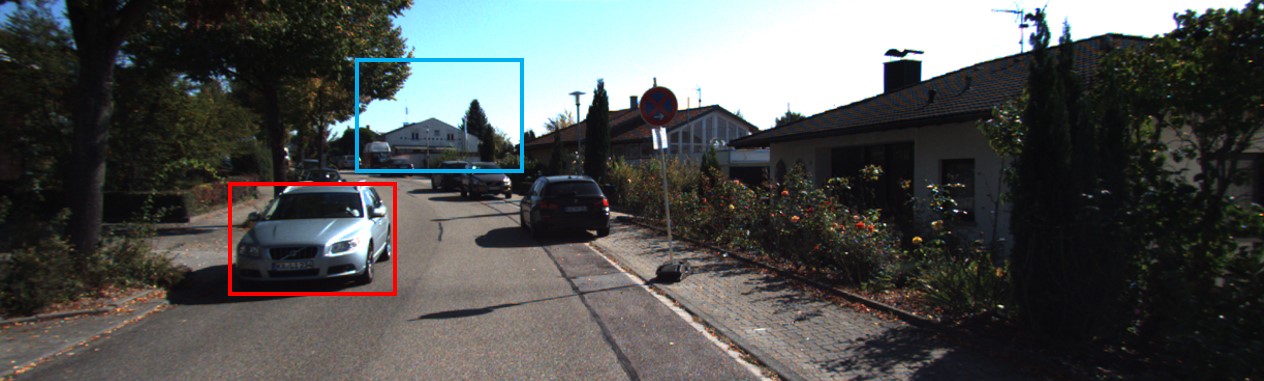}}} & 
     \multicolumn{3}{c}{\makecell{\includegraphics[width=.30\linewidth]{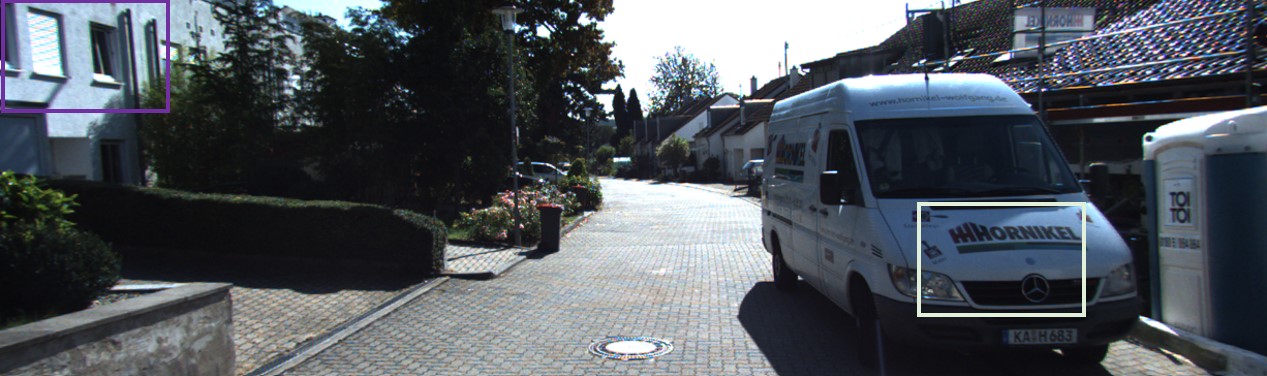}}} & 
     \multicolumn{3}{c}{\makecell{\includegraphics[width=.30\linewidth, height=.09\linewidth]{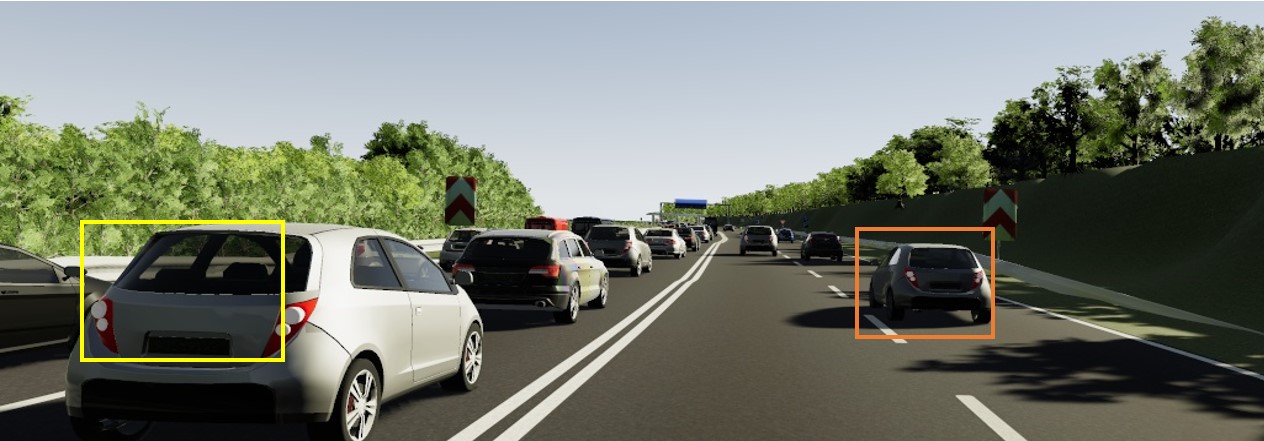}}}
\end{tabular}}

  \resizebox{\textwidth}{!}{
\begin{tabular}{ccccccccc}  

     \hspace{5.5pt}
     \makecell{\includegraphics[width=.0975\linewidth]{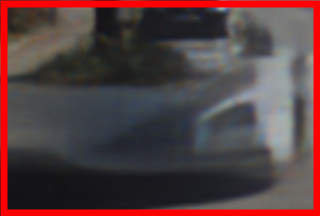}} & 
     \makecell{\includegraphics[width=.0975\linewidth]{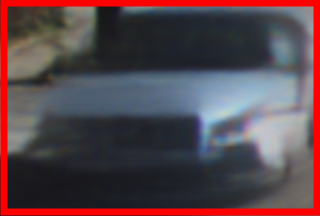}} & 
     \makecell{\includegraphics[width=.0975\linewidth]{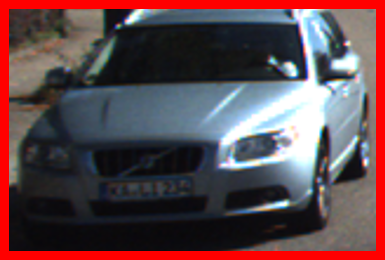}}&
     \makecell{\includegraphics[width=.0975\linewidth]{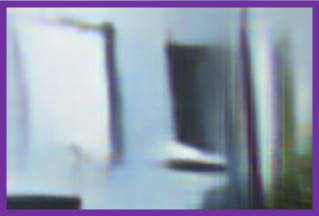}} & 
     \makecell{\includegraphics[width=.0975\linewidth]{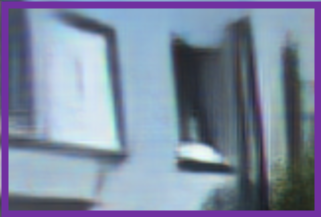}}& 
     \makecell{\includegraphics[width=.0975\linewidth]{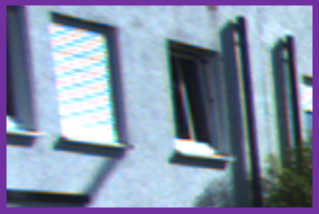}}&

      \makecell{\includegraphics[width=.0975\linewidth]{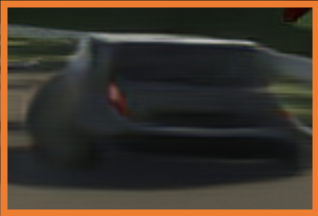}} & 
     \makecell{\includegraphics[width=.0975\linewidth]{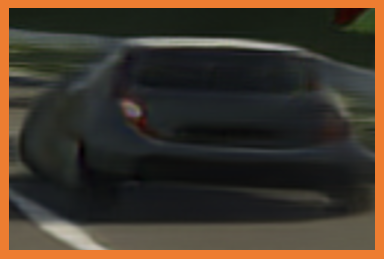}}& 
     \makecell{\includegraphics[height=1.205cm, width=.0975\linewidth]{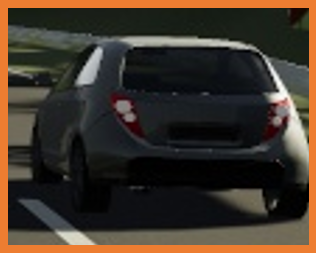}}\\

     \hspace{5.5pt}
     \makecell{\includegraphics[width=.0975\linewidth]{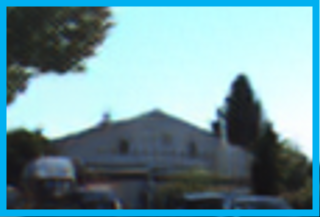}} & 
     \makecell{\includegraphics[width=.0975\linewidth]{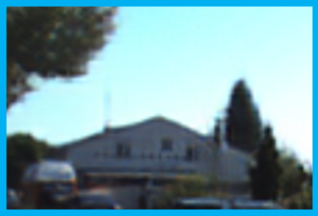}} & 
     \makecell{\includegraphics[width=.0975\linewidth]{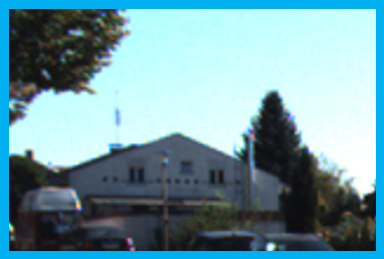}}&
     \makecell{\includegraphics[width=.0975\linewidth]{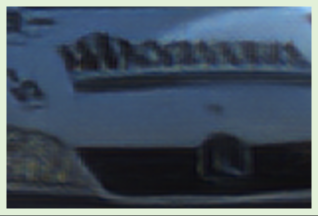}}& 
     \makecell{\includegraphics[width=.0975\linewidth]{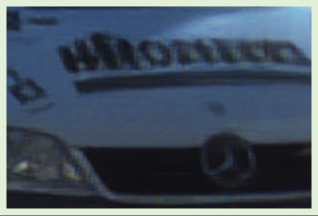}}& 
     \makecell{\includegraphics[width=.0975\linewidth]{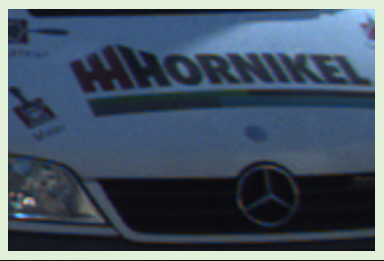}}&
     \makecell{\includegraphics[width=.0975\linewidth]{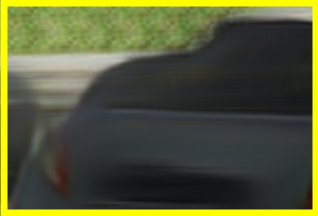}}& 
     \makecell{\includegraphics[width=.0975\linewidth]{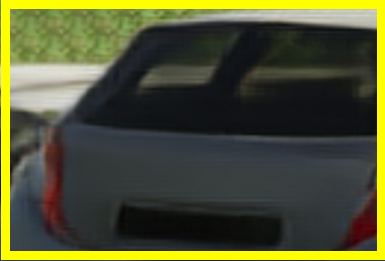}}& 
     \makecell{\includegraphics[height=1.205cm, width=.0975\linewidth]{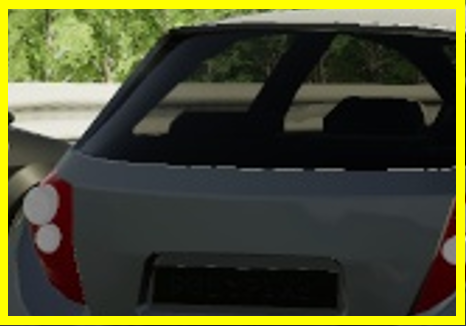}}

  \end{tabular}}
    \caption{\textbf{The Qualitative Comparison of Novel View Synthesis at the Urban Streets and Highway Scenes on KITTI and VKITTI2 Benchmarks.} In the cases of vehicles (Red, Orange, and Yellow), buildings (Light Blue and Purple), text signs (Light Green), our \FrameworkNM{} framework generates more realistic results by estimating more superior poses than PVO. In the bottom rows, The three cropped boxes with same color are corresponding to the rendered areas of PVO, Ours, and Ground Truth (GT), respectively. \textbf{(Best viewed by zoom-in.)}}
    \label{fig:SR_OUTPUT}
\end{figure*}

\begin{figure*}[t!]\scriptsize
  \centering
  \setlength{\tabcolsep}{1pt}
  \resizebox{\textwidth}{!}{
  \begin{tabular}{cccccc}
     
      \makecell{\rotatebox{90}{PVO \cite{ye2023pvo}}}&{\makecell{\includegraphics[width=.16\linewidth]{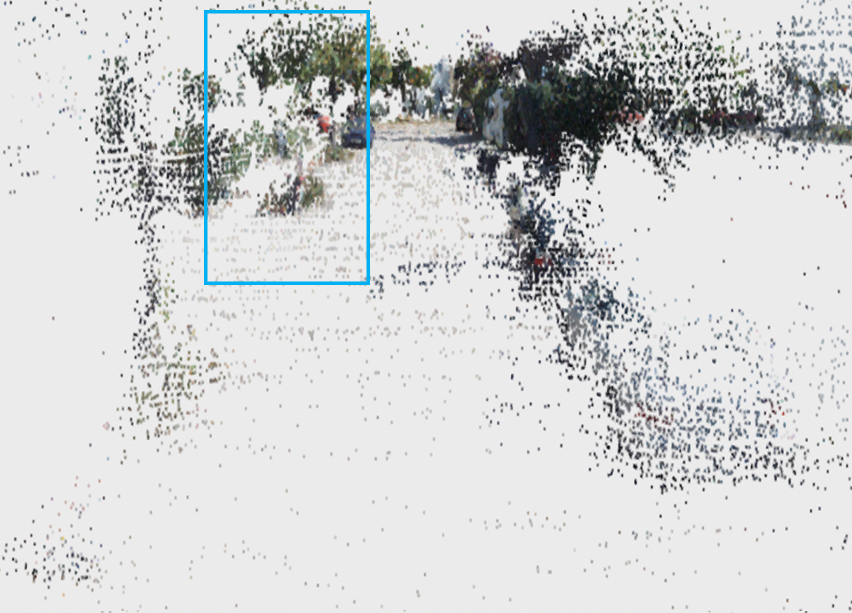}}} & 
     \multicolumn{1}{c}{\makecell{\includegraphics[width=.16\linewidth]{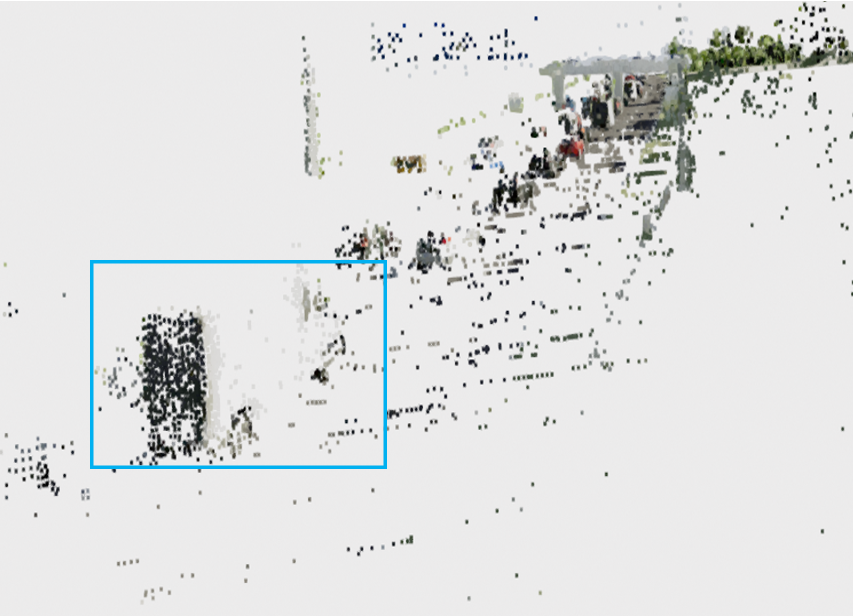}}}&
     \multicolumn{1}{c}{\makecell{\includegraphics[width=.16\linewidth]{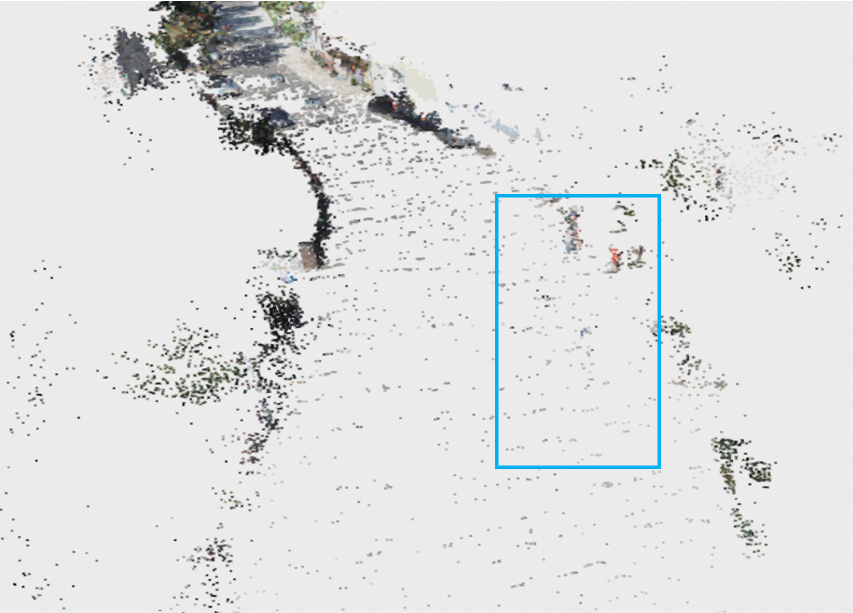}}}&
     \multicolumn{1}{c}{\makecell{\includegraphics[width=.16\linewidth]{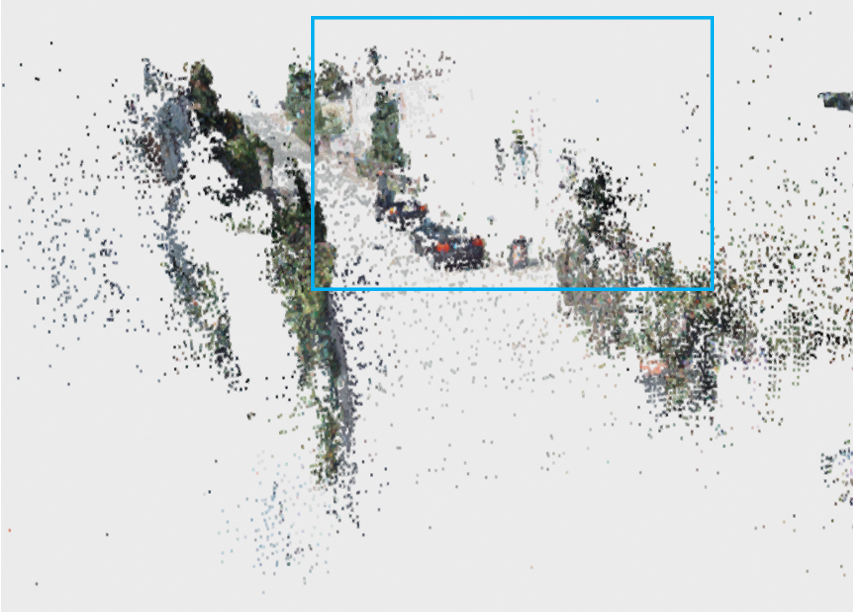}}}&
     \multicolumn{1}{c}{\makecell{\includegraphics[width=.16\linewidth]{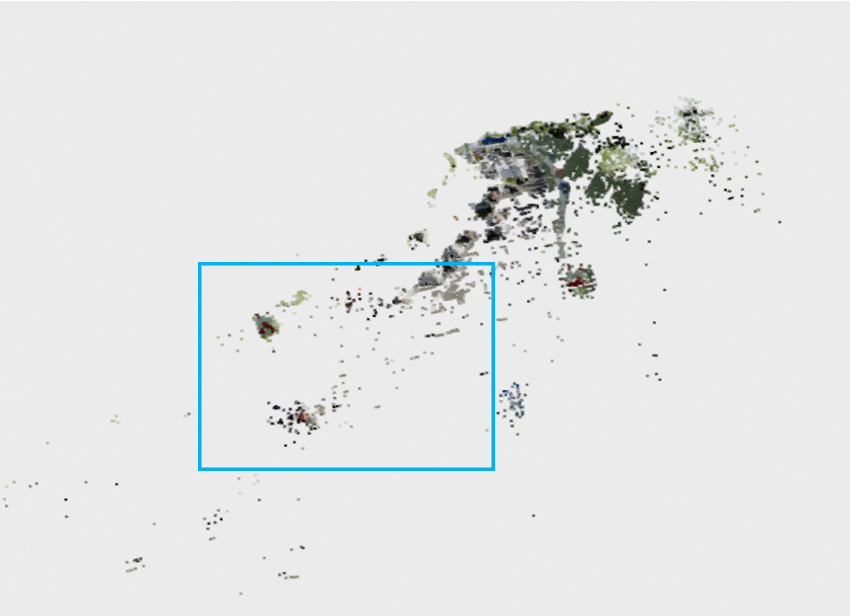}}}\\
     
   \makecell{\rotatebox{90}{Ours}}&{\makecell{\includegraphics[width=.16\linewidth]{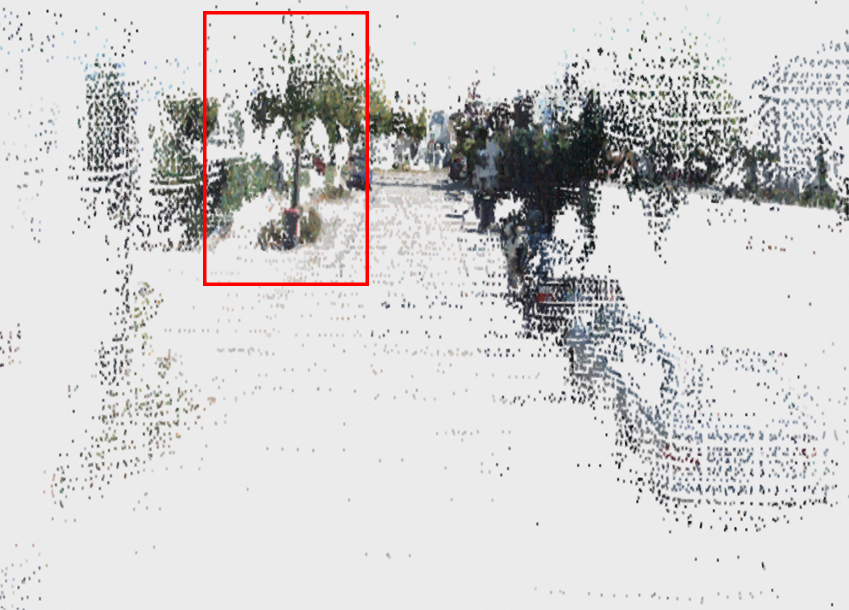}}} & 
     \multicolumn{1}{c}{\makecell{\includegraphics[width=.16\linewidth]{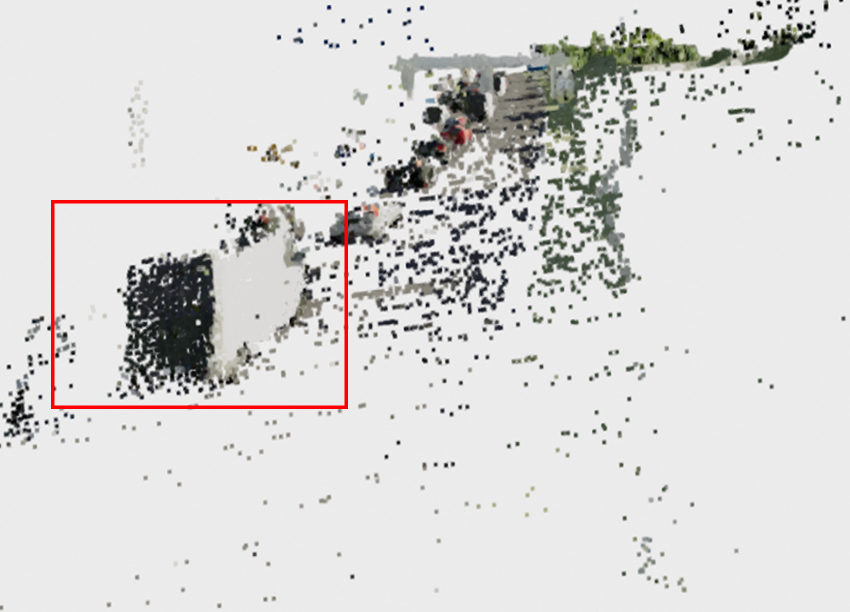}}}&
     \multicolumn{1}{c}{\makecell{\includegraphics[width=.16\linewidth]{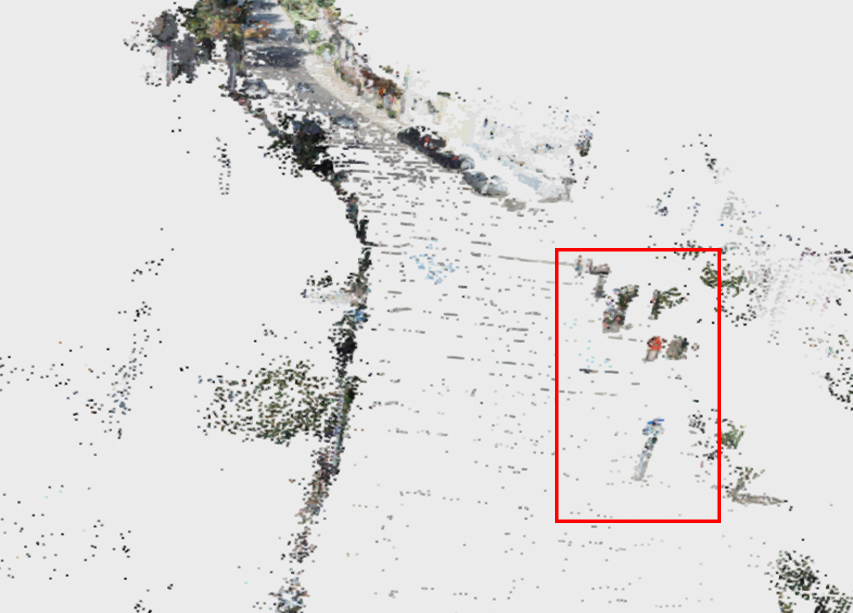}}}&
     \multicolumn{1}{c}{\makecell{\includegraphics[width=.16\linewidth]{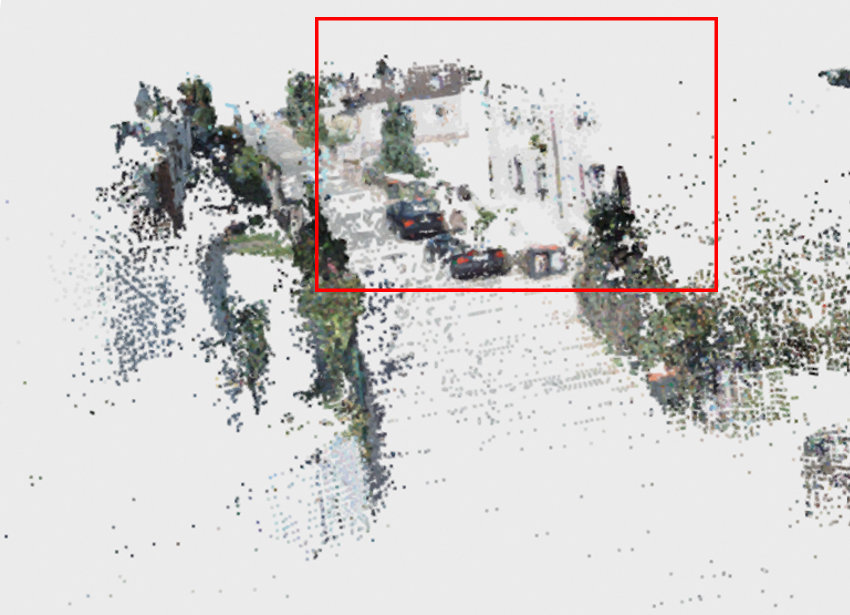}}}&
     \multicolumn{1}{c}{\makecell{\includegraphics[width=.16\linewidth]{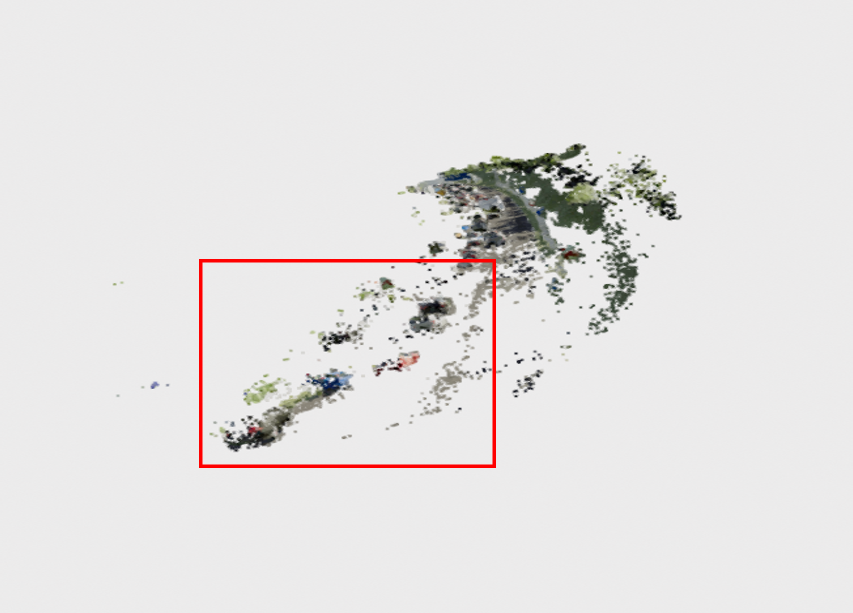}}}
     
\end{tabular}}
    \caption{\textbf{Visualization of Point Cloud from Scene Rendering of Our Method and PVO.} Our method generate denser and more complete point cloud, especially for moving objects, as highlighted by the red boxes. More complete point cloud is beneficial to render precise views. (Best viewed by zoom-in.)}
    \label{fig:PC_OUTPUT}
\end{figure*}

As shown in Table \ref{table1_SR}, we conduct experiments on three distinctive sequences, the KITTI sequence 09, the KITTI sequence 10, and the VKITTI2 sequence 20. It shows that with more accurate camera poses and point cloud positions, more realistic scenes can be rendered. The visualization results from different VO poses and cloud points are illustrated in Fig. \ref{fig:SR_OUTPUT}. To demonstrate the effectiveness of our depth-propagation strategy in VO, the point cloud with different density is shown in Fig. \ref{fig:PC_OUTPUT}.

In order to compare our \FrameworkNM{} with more scene rendering methods, we report view synthesis results in Table \ref{table2_SR} on all sequences of VKITTI2, It is observed that the pose generated by our \FrameworkNM{} is better than the methods like no-pose estimation and Colmap initialization. Our methods handles dynamic scene rendering better than the NeRF + Time.

\begin{table}[t!]
\centering
\caption{\textbf{The Comparison of Scene Rendering with Other Methods.} The bold represents the best, and the underlined represents the second.}
\label{table2_SR}

\begin{tabular}{c| c c c}
\hline 
\multirow{2}{*}{} & \multicolumn{3}{c}{VKITTI2} \\
\hline
Method \multirow{2}{*}{}& PSNR↑ & LPIPS↓ & SSIM↑ \\
\hline
NeRF\cite{mildenhall2021nerf}  &18.67 &0.634 &0.548\\ 

NeRF\cite{mildenhall2021nerf}+Time  &19.03 &0.587 &0.574\\ 

iMAP\cite{sucar2021imap}  &20.02 &- &-\\

NICE-SLAM\cite{zhu2022nice} &19.15 &- &-\\

Mip-NeRF\cite{barron2021mip} &19.91 &- &-\\

Instant-NGP\cite{muller2022instant} &21.58 &- &0.630\\

Splatam (masked) \cite{keetha2024splatam} &20.00 &- &\textbf{0.740}\\

NSG\cite{ost2021neural} &\textbf{23.41} &\underline{0.317} &0.689\\ 

Ours  &\underline{22.39} &\textbf{0.199} &\underline{0.727}\\ 
\hline 
\end{tabular}
\end{table}

After learning neural descriptors from sparse point clouds, our model is capable to synthesize novel views for urban scenes. In Fig. \ref{NVS_ROTATE}, we further render several novel views by rotating the observation directions in a fixed spot, to demonstrate the effectiveness of our neural rendering approach and its potential applications to boost the simulation efficiency in the view-limited driving scenes.

\subsection{Ablation Study}
\label{sec::ablation}
In this section, we verify the necessities of our updated strategies through four ablation study experiments.
\begin{enumerate}

\begin{table*}[b!]
\centering
\caption{\textbf{The Ablation Study Performance of Different Video Panoptic Segmentation Models on the VKITTI2}.}
\resizebox{\textwidth}{!}{
\begin{tabular}{c| c| c| c| c| c}
\hline
\multirow{2}{*}{Methods} & \multicolumn{4}{c|}{VPQ with Different Temporal Window Size} & \multirow{2}{*}{VPQ Mean}\\
\cline{2-5}

\multirow{2}{*}{} & k=0 & k=5 & k=10 & k=15 & \\

\hline
Ours (VO\(\to\)VPS) & 60.8 / 57.7 / 61.6 & 56.0 / 45.6 / 58.9 & 53.6 / 37.9 / 57.9 & 52.0 / 32.6 /57.3 & 55.5 / 43.3 / 58.9\\

Ours (VO\(\to\)VPS $\times$ 2) &60.8 / 57.7 / 61.7 & 56.1 / 45.1 / 59.2 & 53.6 / 37.0 / 58.2 & 52.2 / 31.7 /57.8 & 55.7 / 42.9 / 59.2\\

Ours (VO\(\to\)VPS $\times$ 3) &\textbf{60.8} / 57.7 / 61.7 & \textbf{56.1} / 44.9 / 59.3 & \textbf{53.7} / 36.9 / 58.4 & \textbf{52.3} / 31.5 /58.0 & \textbf{55.7} / 42.7 / 59.3\\
\hline

Ours w/ Vanilla Fusion \cite{ye2023pvo} & 60.5 / 56.1 / 61.7 & 55.4 / 43.2 / 58.7 & 52.5 / 34.7 / 27.3 & 50.8 / 29.1 /56.8 & 54.8 / 40.7 / 58.6\\
\hline

Ours w/ AG Fusion + SE \cite{hu2018squeeze}  &60.7 / 57.2 / 61.7 & 56.6 / 45.6 / 59.5 & 53.9 / 37.6 / 58.4 & 52.6 / 32.5 / 58.1 & 55.9 / 43.2 / 59.4\\

Ours w/ AG Fusion + CBAM \cite{woo2018cbam} &60.8 / 57.0 / 61.8 & 56.5 / 45.4 / 59.6 & 54.0 / 38.3 / 58.2 & 52.7 / 33.4 / 58.0 & 56.0 / 43.5 / 59.4\\

Ours w/ AG Fusion + ECA \cite{wang2020eca}  & 60.8 / 57.5 / 61.7 & 56.8 / 47.2 / 59.4 & 54.4 / 40.3 / 58.3 & 53.3 / 35.6 / 58.1 & 56.3 / 45.1 / 59.3\\

Ours w/ AG Fusion + EMA \cite{ouyang2023efficient} & \textbf{60.9} / 57.7 / 61.8 & \textbf{57.1} / 48.1 / 59.5 & \textbf{54.7}/ 41.8 / 58.2 & \textbf{53.6} / 37.9 / 57.8 & \textbf{56.6} / 46.4 / 59.3\\

\hline
\end{tabular}
}
\label{table_Ablation study_1}
\end{table*}

\begin{table*}[hb!]
\centering
\caption{\textbf{The Ablation Study Performance of Different AG Fusion kernel sizes on the VIPER.}}
\begin{tabular}{c| c| c| c| c| c}
\hline
\multirow{2}{*}{Methods} & \multicolumn{4}{c|}{VPQ with Different Temporal Window Size} & \multirow{2}{*}{VPQ Mean}\\
\cline{2-5}

\multirow{2}{*}{} & k=0 & k=5 & k=10 & k=15 & \\
\hline

Kernel Size = 3 & 34.0 / 17.3 / 46.9 & 31.1 / 12.1 / 45.7 & 30.0 / 9.4 / 45.7 & 29.6 / 8.0 /46.2 & 31.1 / 11.7 / 46.1\\

Kernel Size = 5 & 34.3 / 17.7 / 47.1 & 31.2 / 12.3 / 45.8 & 30.4 / 10.0 / 46.0 & 30.2 / 8.9 /46.6 & 31.5 / 12.2 / 46.3\\

Kernel Size = 7 & \textbf{34.5} / 17.8 / 47.3 & \textbf{31.5} / 12.7 / 46.0 & \textbf{30.6} / 10.6 / 46.0 & \textbf{30.3} / 9.4 /46.5 & \textbf{31.7} / 12.6 / 46.4\\
\hline

\end{tabular}
\label{table_Ablation study_2}
\end{table*}

    \item \textbf{\textit{Iterative Interaction Mechanisms.}} Through the iterative interaction with the VPS module, the geometric information provided by the VO module can continuously improve the segmentation performance and reduce geometric estimation errors in return. First three rows in Table \ref{table_Ablation study_1} discloses that using VO-estimated geometric information, indicating the more accurate geometric information obtained iteratively, can continuously enhance the segmentation performance. Given more iterations, the slight performance improvement is difficult to be observed, and it can gradually approach the performance of fusion using ground-truth geometric information.
    
    \item \textbf{\textit{Different Attention-based Mechanisms.}} The result of combining our AG Fusion module with different channel attention mechanisms, such as SE \cite{hu2018squeeze}, CBAM \cite{woo2018cbam}, ECA \cite{wang2020eca}, and EMA \cite{ouyang2023efficient} are reported at the last four rows of Table \ref{table_Ablation study_1}, it demonstrates the generalizability of our fusion strategy and its capability to integrate with other cutting-edge attention mechanisms.

\begin{figure}[t!]\footnotesize
  \centering
  \setlength{\tabcolsep}{0.7pt}
  \begin{tabular}{ccc}
     & \multicolumn{2}{c}{{Original Image @ $0^\circ$}}\\
     & \multicolumn{2}{c}{\makecell{\includegraphics[width=.47\linewidth]{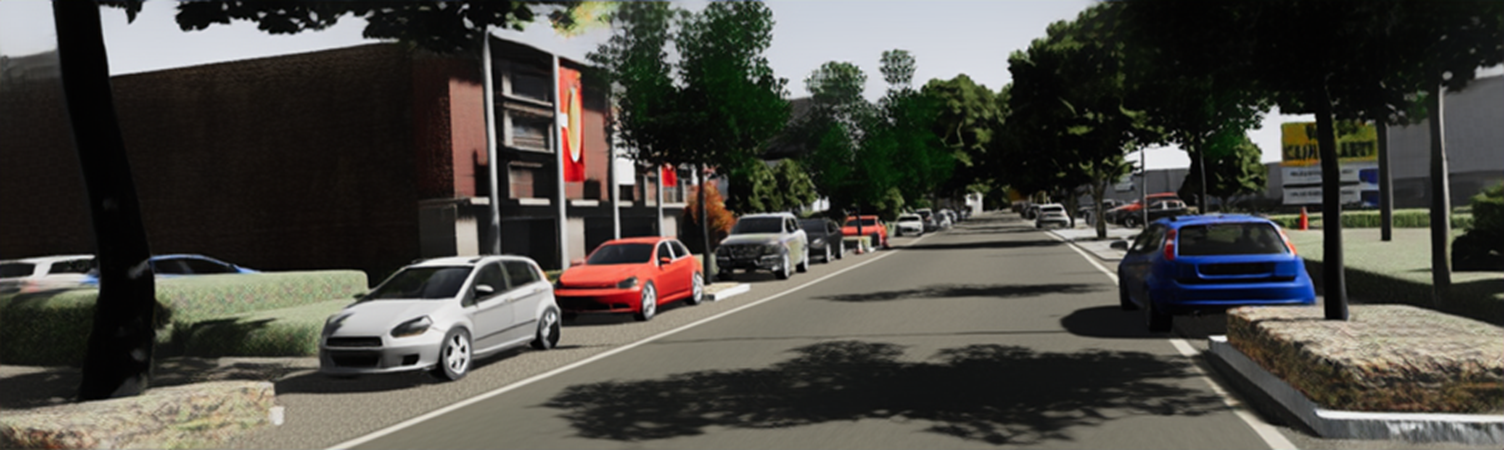}}} \\
     & \multicolumn{2}{c}{Rendered Novel Images @ Different Angles}\\

     \makecell{$1^\circ$}  & \makecell{\includegraphics[width=.47\linewidth]{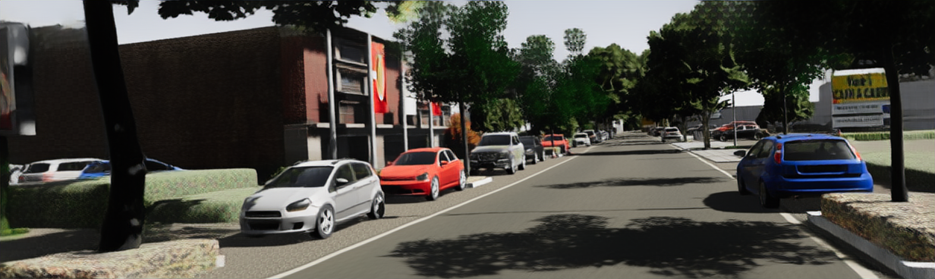}} & 
     \makecell{\includegraphics[width=.47\linewidth]{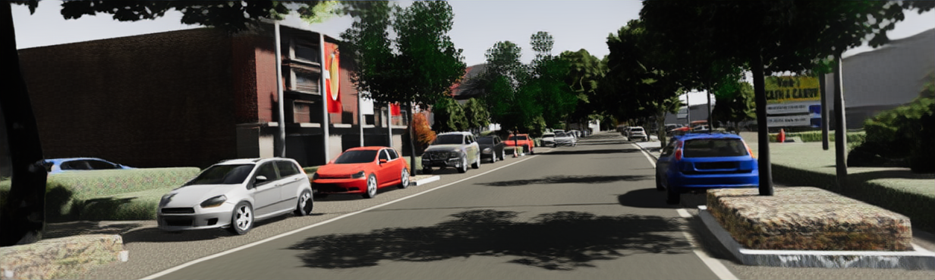}} \\
     \makecell{$2^\circ$}  & \makecell{\includegraphics[width=.47\linewidth]{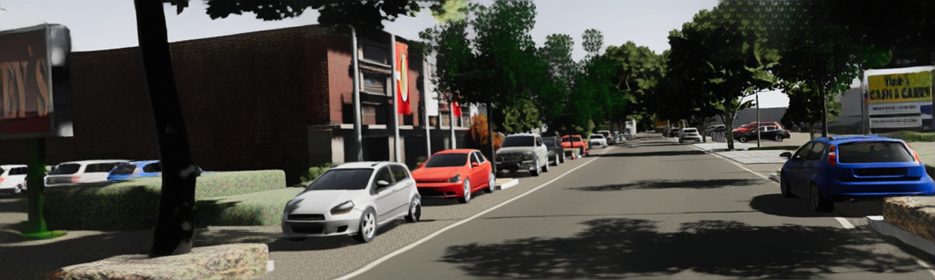}} &
     \makecell{\includegraphics[width=.47\linewidth]{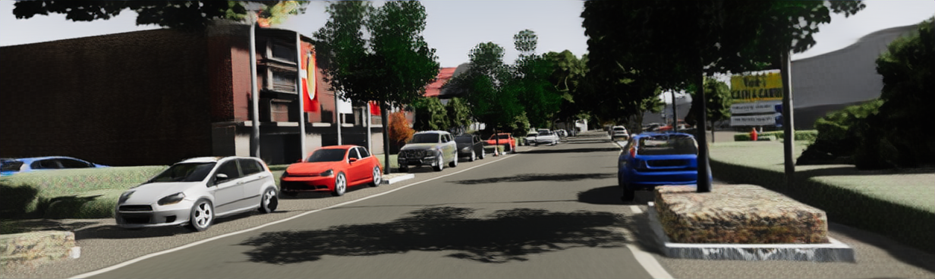}}\\ 
     \makecell{$3^\circ$}  & \makecell{\includegraphics[width=.47\linewidth]{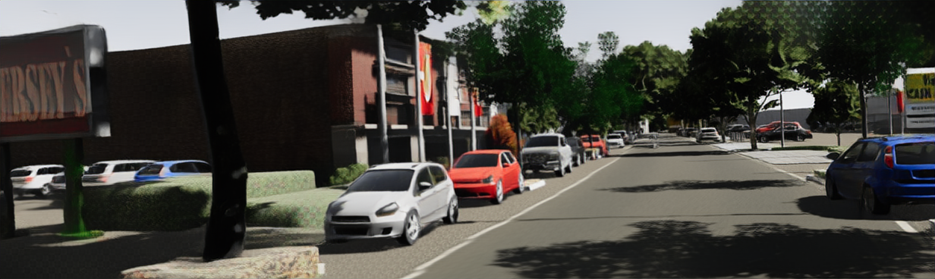}} &
     \makecell{\includegraphics[width=.47\linewidth]{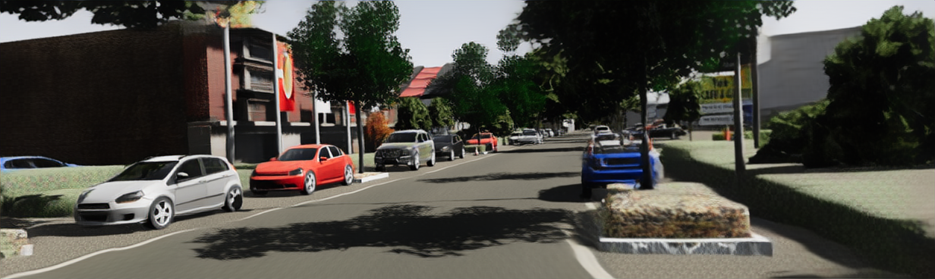}}\\
     \makecell{$4^\circ$}  & \makecell{\includegraphics[width=.47\linewidth]{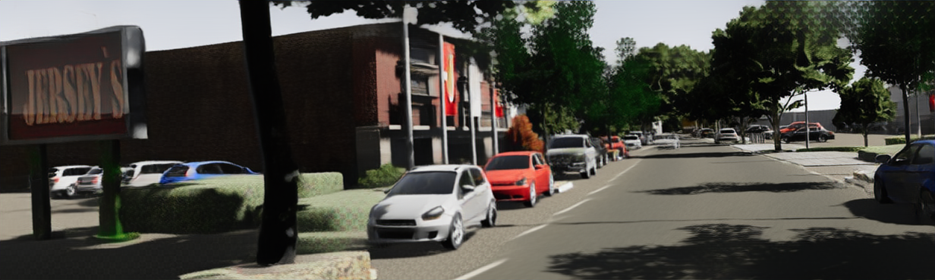}} &
     \makecell{\includegraphics[width=.47\linewidth]{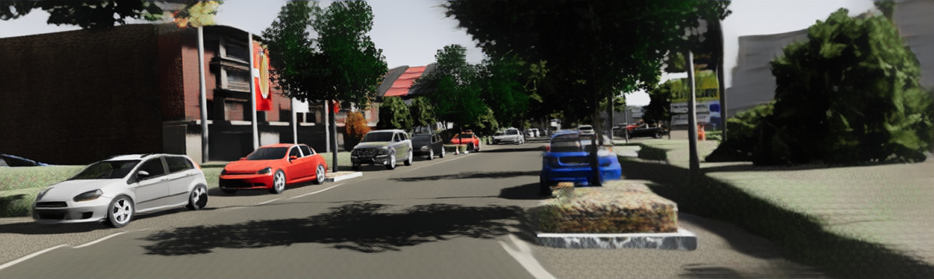}}\\
     \makecell{$5^\circ$}  & \makecell{\includegraphics[width=.47\linewidth]{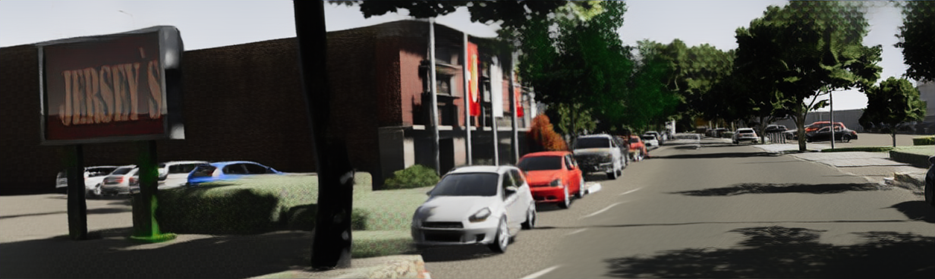}} &
     \makecell{\includegraphics[width=.47\linewidth]{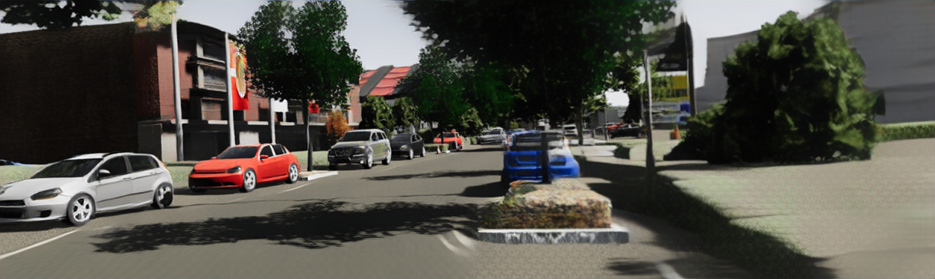}}\\
     \makecell{}  & { Rotate to the Left } & { Rotate to the Right } \\
  \end{tabular}
    \caption{
    \textbf{Demonstration of Novel Views Synthesis Based on the Training via VKITTI2 Data.} Novel views are rendered by rotating the observation directions at a fixed spot, 5 novel views with $1^{\circ}$ step are gradually rendered for each side. The rendering result of left side is more geometric-realistic than the right side, due to the field of views of original image. \textbf{(Best viewed by zooming in.)}
    }
    \label{NVS_ROTATE}
\end{figure}

     \item \textbf{\textit{Fusion Strategies.}} Through the ablation study in the 4th row and the last row of Table \ref{table_Ablation study_1}, we confirm that AG Fusion strategy achieves better segmentation performance compared to the vanilla fusion strategy. AG Fusion strategy integrating the feature maps used by each kernel update layer and adaptively learning fusion weights for different layers, while vanilla fusion solely connecting the feature maps generated by the backbone.
     
    \item \textbf{\textit{Fusion Kernel Sizes.}} 
    We found that the convolution kernel size of the fusion module needs to be adjusted based on the image resolutions. In the VKITTI2 dataset, where the image resolution is $1242 \times 375$, the ideal kernel size is $3 \times 3$. However, when training the VIPER dataset with a resolution of $1920 \times 1080$, it needs to increase the kernel size to $7 \times 7$, to achieve a larger receptive field, for a more robust segmentation results. Table \ref{table_Ablation study_2} reports the VPQ results on the VIPER dataset with different kernel sizes. 
\end{enumerate}  
\section{Conclusion}

This paper presents \FrameworkNM{}, a multi-modal based simultaneous localization and mapping framework designed for the holistic understanding of the urban scenes. The core discovery of our \FrameworkNM{} is that incorporating an attention-based adaptive geometry fusion mechanism is able to boost the segmentation performance, which is reversely beneficial to the odometry and instances tracking tasks. We also introduces a point-based rendering method to render photo-realistic scenes based on the high-quality pose estimation of odometry. Our \FrameworkNM{} is the first work to be evaluated across multiple datasets, to validate that optical flows, poses, and depth information can effectively enhance the vision-based scenes understanding. One drawback for our current work is the strict requirement to the dataset, which impedes the \FrameworkNM{} to be compared with more related methods. Our \FrameworkNM{} manifests that digging out the mutual benefits of stacking multiple tasks, such as segmentation, tracking, localization, and rendering, into a unified framework is a promising direction to be explored, and we plan to enroll the 3D dense mapping task in to our framework in the future work.





\bibliographystyle{IEEEtran}
\bibliography{IEEEabrv,10_MYBIB}

 
%



\section{Biography Section}
 

\vspace{-70pt}
\begin{IEEEbiography}[{\includegraphics[width=1in,height=1.25in,clip,keepaspectratio]{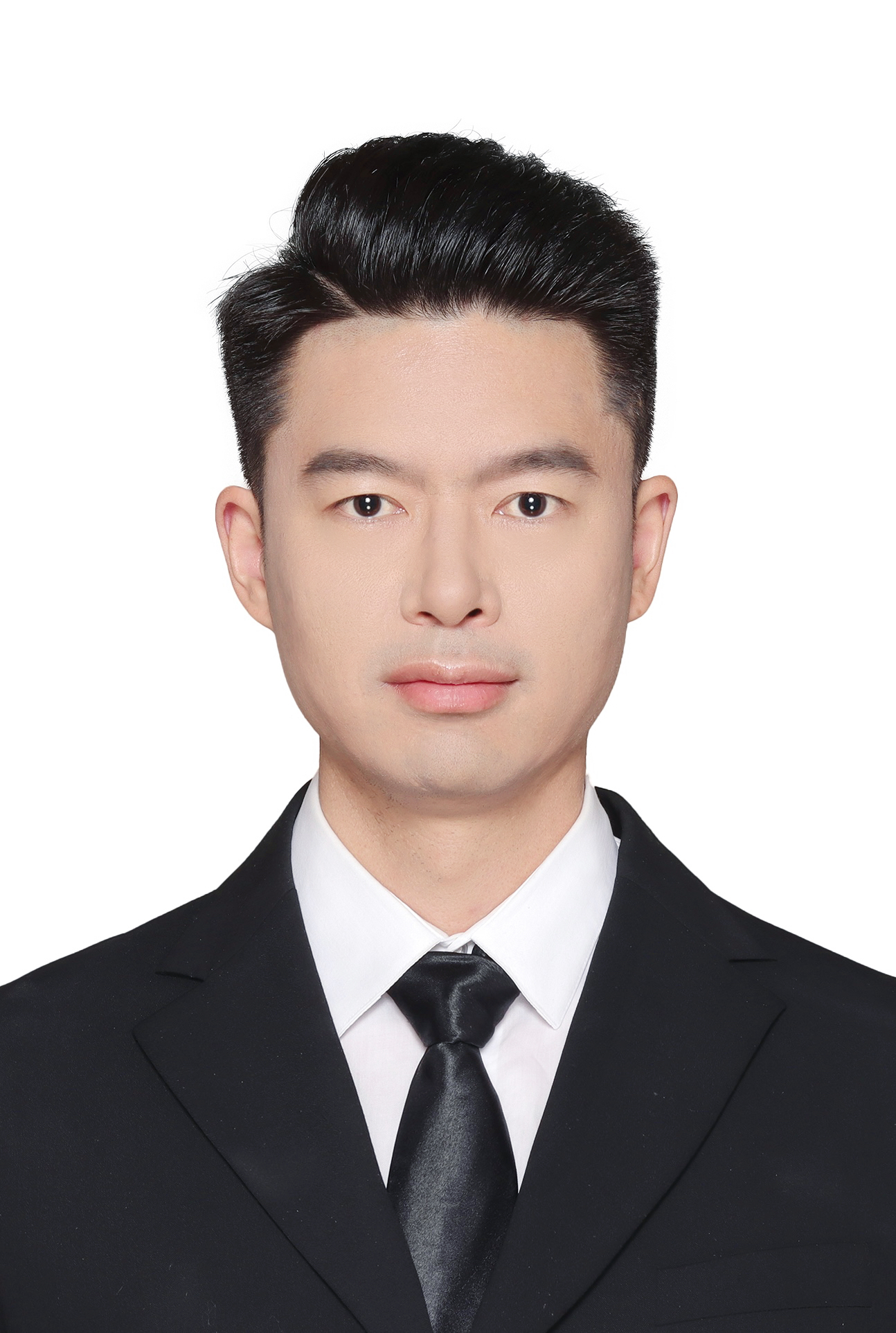}}]{Zhiliu Yang} received his Ph.D. in Electrical and Computer Engineering from Clarkson University, Potsdam, New York, USA, in 2021. Currently he is an assistant professor in the School of Information Science and Engineering at Yunnan University, Kunming, China. He was a recipient of Young Talents Award of Xing Dian Program of Yunnan Province, China. His current research interest is in machine vision, embodied AI, scene mapping and generation, dynamic scene understanding, and unified modeling for multi-task learning.
\end{IEEEbiography}
\vspace{-20pt}

\begin{IEEEbiography}[{\includegraphics[width=1in,height=1.25in,clip,keepaspectratio]{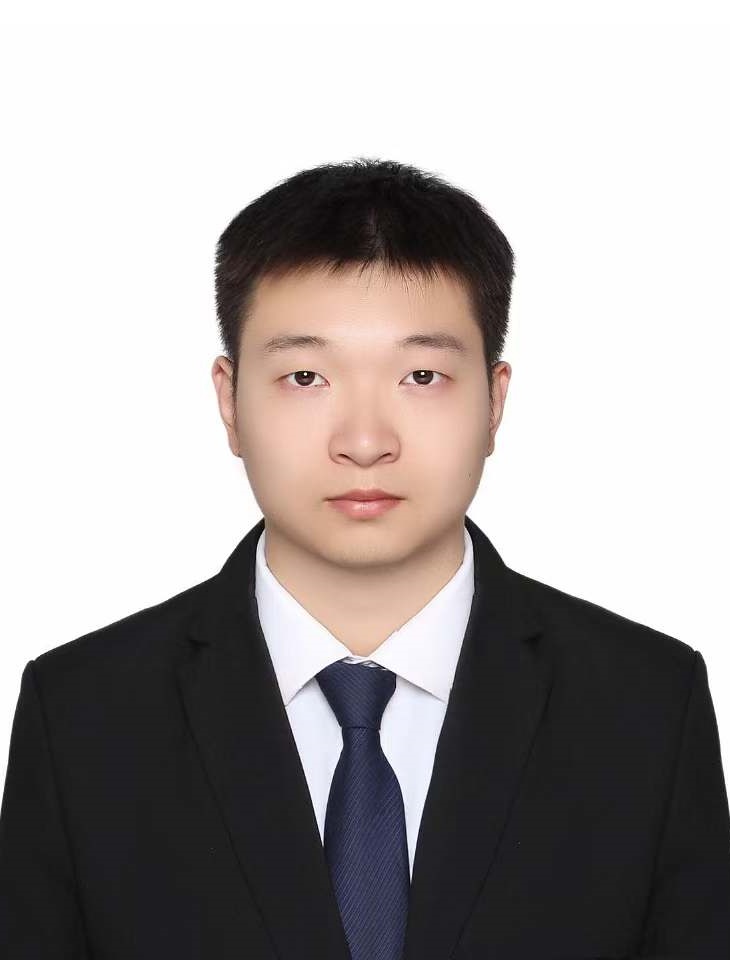}}]{Jinyu Dai} received his B.E. degree in Communication Engineering from Kunming University of Science and Technology (KUST), Kunming, China, in 2022. He is currently working toward the M.S. degree with the School of Information Science and Engineering, Yunnan University, Kunming, China. His research interests include visual perception and scene reconstruction.
\end{IEEEbiography}
\vspace{-20pt}

\begin{IEEEbiography}[{\includegraphics[width=1in,height=1.25in,clip,keepaspectratio]{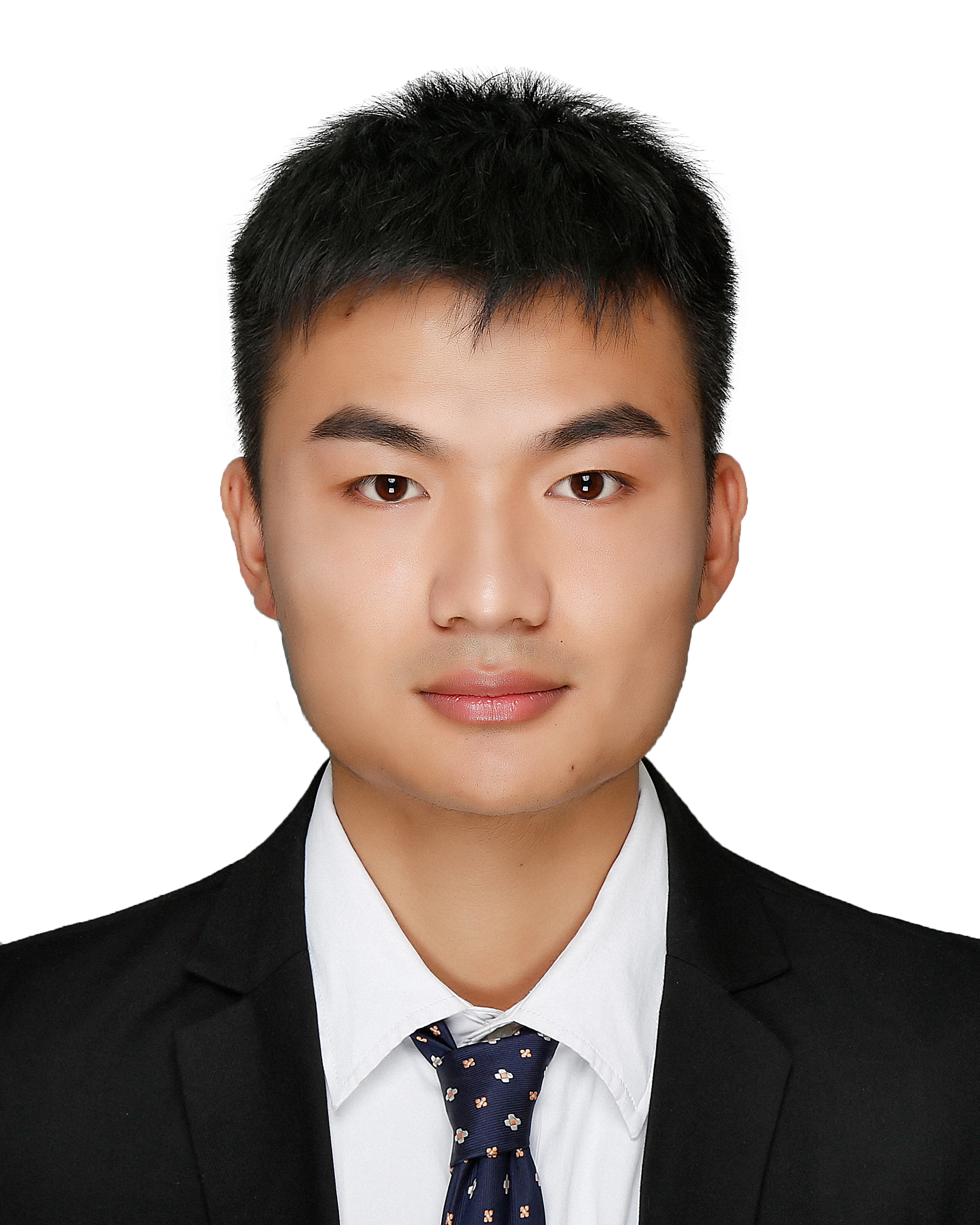}}]{Jianyuan Zhang} received his B.E. degree in Electronics Science and Technology from the Yunnan University, Kunming, China, in 2020. He is currently working toward the M.S. degree with the School of Information Science and Engineering, Yunnan University, Kunming, China. His research interests include visual perception, LiDAR odometry and implicit scene reconstruction.
\end{IEEEbiography}
\vspace{-20pt}

\begin{IEEEbiography}[{\includegraphics[width=1in,height=1.25in,clip,keepaspectratio]{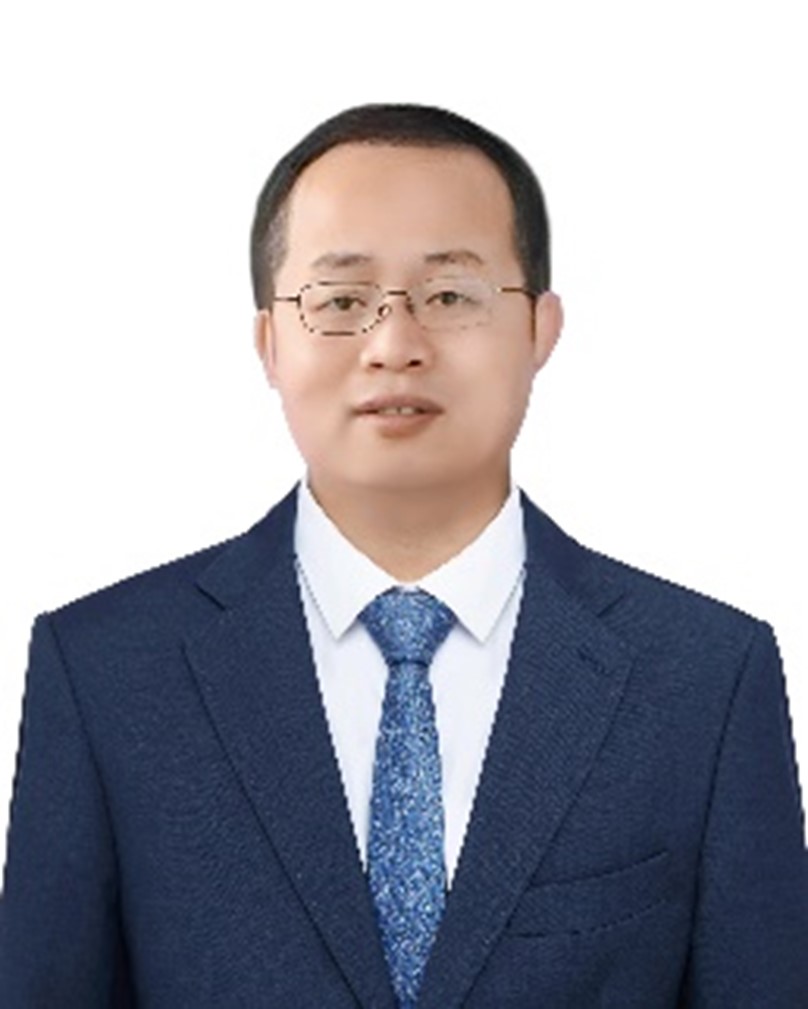}}]{Zhu Yang} received his Ph.D. degree from Beijing Institute of Technology in 2019. Currently, he is an assistant professor in the School of Information and Electronics, Beijing Institute of Technology. His research interests are chip design and signal processing related to intelligent remote sensing.
\end{IEEEbiography}





\vfill
\end{document}